\documentclass[journal]{IEEEtran}
 
\makeatletter
\def\ps@headings{%
	\def\@oddhead{\mbox{}\scriptsize\rightmark \hfil \thepage}%
	\def\@evenhead{\scriptsize\thepage \hfil \leftmark\mbox{}}%
	\def\@oddfoot{}%
	\def\@evenfoot{}}
\makeatother 

\makeatletter
\def\endthebibliography{%
	\def\@noitemerr{\@latex@warning{Empty `thebibliography' environment}}%
	\endlist
}
\makeatother

\usepackage{graphicx}
\usepackage{amsmath, amsfonts,epsfig, multirow, floatflt}
\usepackage{amssymb}
\usepackage{subfigure}
\usepackage{cite}
\usepackage{algorithm}
\usepackage{algorithmicx}
\usepackage{algpseudocode}
\usepackage{enumitem}
\usepackage{diagbox}
\usepackage{mathrsfs}
\usepackage{url}
\usepackage{color}

\usepackage{caption}
\usepackage{hhline}
\usepackage{array}
\usepackage{booktabs}
\usepackage{amsthm}
\usepackage{bm}
\usepackage{setspace}
\usepackage[table,xcdraw]{xcolor}
\usepackage{stfloats}

\newcommand{\MYnewpage}{%
	\ifCLASSOPTIONonecolumn
		\ifCLASSOPTIONjournal
			\typeout{The onecolumn journal mode.}
			\newpage
		\fi
	\fi}

\allowdisplaybreaks[4]
\graphicspath{{figure/}}

\begin{document}
\ifCLASSOPTIONonecolumn
	\typeout{The onecolumn mode.}
	\title{Learning Value of Information towards Joint Communication and Control in 6G V2X}
	\author{Lei~Lei,~\IEEEmembership{Senior~Member,~IEEE},  
	}
\else
	\typeout{The twocolumn mode.}
	\title{Learning Value of Information towards Joint Communication and Control in 6G V2X}
	\author{Lei~Lei,~\IEEEmembership{Senior~Member,~IEEE}, Kan~Zheng,~\IEEEmembership{Fellow,~IEEE}, and~Xuemin (Sherman)~Shen,~\IEEEmembership{Fellow,~IEEE}
    
\thanks{Manuscript received December 24, 2024; revised June 26, 2025; accepted August 4, 2025. This work was supported by the Natural
Sciences and Engineering Research Council of Canada (NSERC) under
Discovery Grant. (Corresponding author: Lei Lei.)}

\thanks{L. Lei is with the College of Engineering, University of Guelph, Guelph, Ontario, Canada (e-mail: leil@uoguelph.ca).}

\thanks{K. Zheng is with the College of Electrical Engineering and Computer Sciences, Ningbo University, Ningbo, 315211, China.}

\thanks{X. Shen is with the Department of Electrical and Computer
Engineering, University of Waterloo, Waterloo, Ontario, Canada.}
}

\fi

\maketitle

\ifCLASSOPTIONonecolumn
	\typeout{The onecolumn mode.}
	\vspace*{-50pt}
\else
	\typeout{The twocolumn mode.}
\fi
\begin{abstract}
 
As Cellular Vehicle-to-Everything (C-V2X) evolves towards future sixth-generation (6G) networks, Connected Autonomous Vehicles (CAVs) are emerging to become a key application. Leveraging data-driven Machine Learning (ML), especially Deep Reinforcement Learning (DRL), is expected to significantly enhance CAV decision-making in both vehicle control and V2X communication under uncertainty. These two decision-making processes are closely intertwined, with the value of information (VoI) acting as a crucial bridge between them. In this paper, we introduce Sequential Stochastic Decision Process (SSDP) models to define and assess VoI, demonstrating their application in optimizing communication systems for CAVs. Specifically, we formally define the SSDP model and demonstrate that the MDP model is a special case of it. The SSDP model offers a key advantage by explicitly representing the set of information that can enhance decision-making when available. Furthermore, as current research on VoI remains fragmented, we propose a systematic VoI modeling framework grounded in the MDP, Reinforcement Learning (RL) and Optimal Control theories. We define different categories of VoI and discuss their corresponding estimation methods. Finally, we present a structured approach to leverage the various VoI metrics for optimizing the ``When", ``What", and ``How" to communicate problems. For this purpose, SSDP models are formulated with VoI-associated reward functions derived from VoI-based optimization objectives. While we use a simple vehicle-following control problem to illustrate the proposed methodology, it holds significant potential to facilitate the joint optimization of stochastic, sequential control and communication decisions in a wide range of networked control systems.    
\end{abstract}

\ifCLASSOPTIONonecolumn
	\typeout{The onecolumn mode.}
	\vspace*{-10pt}
\else
	\typeout{The twocolumn mode.}
\fi

\IEEEpeerreviewmaketitle

\MYnewpage

    \section{Introduction}

Connected Autonomous Vehicles (CAVs) integrate automated control with advanced communication technologies to achieve self-driving capabilities with minimal or no human intervention.
Cellular Vehicle-to-Everything (C-V2X) technologies are advancing rapidly and are on the brink of commercial deployment. Unlike Autonomous Vehicles (AVs) without communication capabilities, CAVs can achieve a more accurate and comprehensive perception of the driving environment by leveraging critical information exchanged through Vehicle-to-Everything (V2X) communications, which is referred to as `V2X information' in this paper. This enables them to make more reliable and efficient driving decisions by integrating V2X information with local sensory data \cite{6GDR,ACDV2025,10505023}. \par

As C-V2X evolves towards sixth-generation (6G) networks to support CAVs, several key challenges emerge, including Artificial Intelligent (AI)-native architectures, integrated sensing and communication, and increasingly stringent Ultra-Reliable Low-Latency Communication (URLLC) requirements\cite{CAV6G}. Among these, a fundamental requirement is the timely and efficient utilization of limited communication resources to transmit critical information for autonomous driving. Unlike traditional networks that focus primarily on raw data transmission, 6G envisions a tight integration of communication and control functions in CAVs, enabling goal-oriented communication that prioritizes task-relevant information. This paradigm shift demands communication systems that are not only fast and reliable but also control-aware, capable of identifying and delivering critical information essential for real-time driving decisions.    \par

Recent years have witnessed a growing interest in applying data-driven Machine Learning (ML) to CAVs. Deep Reinforcement Learning (DRL), which combines deep learning (DL) with reinforcement learning (RL), has shown significant advantages over traditional model-driven control and optimization methods, as it requires fewer restrictive assumptions about the stochastic properties of system dynamics \cite{9351818,9951132,9046279,9119487,Peng2025}. Leveraging DRL for both vehicle control and V2X communication, an intelligent, integrated decision-making system presents a promising solution for ensuring safe and efficient autonomous driving (AD). To enable this, Markov Decision Process (MDP) models must be conceived for framing the autonomous driving (AD) and V2X communication problems, respectively. The system would be comprised of two types of agents - a control agent and a communication agent - that employ DRL algorithms to solve the MDP models and learn optimal policies for decision-making. Notably, the control and communication decisions may operate at different time scales \cite{10376448}.  \par

The control agent makes AD decisions based on available sensory and V2X information. These decisions can occur at various levels, including perception understanding, behavior and motion planning, vehicle control, or an end-to-end approach that encompasses all levels \cite{zhao2024survey}. In this paper, we focus on vehicle control decisions to illustrate our proposed methodology, although it is applicable to all types of AD decisions. The vehicle control decisions made by the control agent — referred to as control inputs — are executed by actuators such as the steering system, throttle, brakes, and gear shifter, thereby enabling autonomous vehicle operation. It is widely recognized that providing the control agent with more information at the time of decision-making can lead to better decisions. However, an overwhelming amount of information can be toxic to the control agent and greatly increases the computational complexity of making decisions due to the curse of dimensionality problem \cite{lei2020deep}. Ideally, the control agent would have access to the necessary and sufficient information to make optimal decisions. In practice, however, information comes at a cost and can be imperfect. This paper mainly focuses on the communication costs incurred for information dissemination and the imperfections in information resulting from non-ideal communication. \par

Given the challenges of perfectly delivering all available V2X information to CAVs, a series of key questions arise: \emph{\textbf{what} information should be transmitted, \textbf{when} should it be transmitted, and \textbf{how} should the limited radio resources be scheduled for transmission?} Thus, the communication agent is responsible for making these decisions.   \par

Understanding the importance of sharing specific information with other CAVs is crucial for making effective communication decisions. To evaluate this, the value of information (VoI) is studied in this paper, as it quantifies the expected benefits of acquiring additional information for decision-making \cite{4082064}. Although VoI has been investigated across diverse research domains, each with its unique perspective \cite{9751706}, this paper specifically focuses on VoI in the context of integrated control and communication. \par

Since the information is utilized by control agents for decision-making, the VoI should be evaluated by these agents. The communication agents - whether at the base station (BS), roadside infrastructure, CAVs and other information sources - can leverage the estimated VoI, alongside traditional communications performance metrics such as throughput and delay, to optimize their communication decisions~\cite{VDTN2024}. \par 
   
The VoI depends on how the control agent reacts to missing or imperfect information by adjusting its decisions, which in turn relies on the type of control algorithm employed. Traditionally, AV control algorithms have been designed based on classical control theory, such as linear controller, $\mathcal{H}_\infty$ controller, and Sliding Mode Controller (SMC) \cite{7990611}. In this paper, we focus on VoI when DRL-based AV control algorithms are employed. However, our results are not limited to DRL and can be extended to any control algorithm formulated through Markov Decision Processes (MDPs), including both model-based dynamic programming and optimal control methods, as well as model-free reinforcement learning approaches.  \par

The MDP model describes a dynamic system whose `state' evolves over time under the influence of `actions' taken at discrete instances of time. The outcome of each action is captured by a scalar reward. The objective is to derive a policy function that maps states to actions, enabling an agent to make a sequence of decisions (on actions) over time that maximizes the expected cumulative rewards. \par 

In MDP formalism, the `state' is a set of variables that contain essential information for decision-making. However, there is a lack of thorough discussion regarding the formal definition of state in the existing literature. Specifically, what information should it incorporate? Considering the communication costs and limitations for information dissemination, the above question can be expanded to the following three questions:

\begin{description}
	\item[-Q1] What information can enhance decision-making if made available?
      \item [-Q2] How can the benefit of information be measured? In other words, how can VoI be defined and estimated?
        \item [-Q3] How can VoI be leveraged to make communication decisions for information dissemination and radio resource management (RRM)?
\end{description}

\subsection{What information can enhance decision-making if made available?}
\subsubsection{Related Works}
    The only requirement for a state in an MDP model is that it satisfies the Markov property. In other words, given the current state and action, the reward and the next state are independent of past states and actions. Silver points out that there are actually two versions of state, namely the environment state and the agent state \cite{silver}. The environment state is the environment's private state representation, which is usually not fully visible to the agent. It includes the information the environment uses to determine the next state and reward, while some of this information may be irrelevant to decision-making. Meanwhile, the agent's policy is a function of the agent state, which consists of the information the agent uses to select the action at the time of decision-making. This state is the agent's internal representation of the environment state. Since the information in the environment state may not be fully and accurately accessible to the agent, the agent state is constructed based on the information available to it. If the agent state does not satisfy the Markov property, the resulting decision process becomes a partial observable MDP (POMDP), which can be reformulated into an MDP by incorporating the history of past states and actions into the agent state.\par 

    Although Silver’s definition distinguishes between the environment state and the agent state, it remains unclear which information in the environment state is irrelevant to decision-making. More importantly, it is not well-defined what information could enhance decision-making if incorporated into the agent state. In contrast, Powell provides a formal definition of the state \cite{Powe11}: ``A state variable is the minimally dimensioned function of history that is necessary and sufficient to compute the decision (policy) function, the transition function, and the contribution (reward) function." He further explains that the definition provides a quick test for the validity of a state variable: ``If there is a piece of data in either the policy function, the transition function, or the reward function that is not in the state variable, then we do not have a complete state variable." However, although the transition and reward functions are objective models of the environment, the policy function is subjective and defined by the agent, with the state as its input argument. This leads to a potential deadlock in using Powell's definition to identify a state: we need to determine the policy function to define the state, yet we also need to ascertain the state to establish the policy function.

\subsubsection{Contribution 1}
    We introduce a general model known as the Sequential Stochastic Decision Process (SSDP) model, of which the MDP model is a special case. Based on the SSDP model, we provide a mathematically rigorous and practical method to identify the set of information that can enhance decision-making if made available.  Specifically, we define exogenous information within the SSDP model as any information that affects the transition function, the reward, or both, but is unavailable prior to decision-making. We demonstrate that this exogenous information can improve decision-making once it becomes available. Moreover, if the exogenous information is not available, any information that can predict the exogenous information can also enhance decision-making. Any other information apart from the above is not useful. We also clarify the relationship between our findings and the definitions of state provided by Silver and Powell.

\subsection{How can the benefit of information be measured? In other words, how can VoI be defined and estimated?}
\subsubsection{Related Works}

From the discussion above, we understand that the agent state may contain only partial information about the environment state. While the agent can utilize its historical data to construct a Markov state, this enhanced state may still be insufficient to fully represent the environment state. In this case, how do we measure the benefit of acquiring additional information that exists in the environment state but is not included in the agent state? Moreover, some information in the agent state may be imperfect due to factors such as transmission delay when delivered to the agent. Thus, how do we measure the benefit of knowing perfect information instead of imperfect information? \par

Information theory provides powerful tools for quantifying the amount of uncertainty in the value of a random variable, and has had a profound influence on the design of communication systems. However, as Howard indicates in his seminal work, ``no theory that involves just the probabilities of outcomes without considering their consequences could possibly be adequate in describing the importance of uncertainty to a decision maker" \cite{4082064}. Howard introduces the theory of VoI, and asserts that numerical values can be assigned to the elimination or reduction of any uncertainty. To explain the concept and intuition of VoI, Howard uses a bidding problem as an example, where the expected profit of eliminating any certainty is used as the VoI. However, Howard's work does not provide a rigorous definition or mathematical formula for  calculating VoI. Following his work, various definitions of VoI have been proposed for different applications and objectives. A survey on VoI in Wireless Sensor Network (WSN) and Internet of Things (IoT) by Alawad et al. \cite{9751706} provides a structured literature review on the fragmented works in this area. Despite these contributions, a theoretical foundation for VoI is still lacking, which would allow VoI to be estimated using uniform formulas for different applications.   \par  

\subsubsection{Contribution 2}
We provide a systematic method  for defining and estimating VoI in the context of making a sequence of decisions under uncertainty. This approach has broad practical applications, with autonomous driving serving as a key example. Our definition of the VoI is rooted in MDP and RL theories, and is derived from the SSDP model we develop. \par

We categorize VoI into two types: (1) utility-based VoI, which measures the performance loss in terms of expected cumulative rewards in SSDP models; and (2) information theory-based VoI, which quantifies the VoI based on how much better we can predict the next state and reward. The rationale for defining information theory-based VoI stems from our assertion in Contribution 1 that only information influencing the state transition and reward contributes to enhanced decision-making.  \par 

For utility-based VoI, we extend Howard's VoI concept and define two versions: Value of Missing Information (VoMI) and Value of Imperfect Information (VoPI). Additionally, we differentiate between Expected Cumulative VoI (EVoI) and Immediate VoI (IVoI) to address the cases when a sequence of decisions are made based on the missing/imperfect information versus when only a single decision is made based on this information. The mathematical relationship between EVoI and IVoI is also established. \par 

To derive the information theory-based VoI, we calculate the conditional Kullback–Leibler (KL) divergence between the probability distribution given by the product of the transition models of the original state and the V2X information, and the probability distribution given by the transition model of the augmented state, which inclues both the original state and the V2X information. While information theory-based VoI is easier to derive given the model of the environment, utility-based VoI has several important advantages. Therefore, we introduce a DRL-based method to estimate the utility-based VoI by offline training policies with missing/imperfect and full/perfect information, respectively. Then, the regret and advantage function can be estimated to quantify the EVoI and IVoI, respectively.\par      

\subsection{How to leverage the VoI to make communication decisions for information dissemination and RRM?}
\subsubsection{Related Works}
VoI has been used to guide communication decisions in existing studies \cite{9815085,8693978,10494374,8887791}. The general idea is to transmit only information with high VoI or prioritizing the transmission of such information. However, VoI in these studies is typically calculated using intuitive metrics, such as the importance of information and its timeliness. Therefore, the vague definition of VoI may lead to misinformed communication decisions. \par

In V2X communication, there is a growing interest aimed at designing RRM mechanisms to minimize the Age of Information (AoI) \cite{9380899,10286022,9460780,9109636}, rather than focusing solely on traditional metrics such as throughput and delay.  While AoI measures the importance of information by assessing its timeliness, it does not account for the broader benefits that information can provide. In contrast, VoI directly quantifies the advantage of information in enhancing the performance of specific tasks. Unfortunately, there is currently a lack of studies focused on VoI-oriented RRM.\par

In MDP and RL communities, the problems of ``What to Communicate" and ``When to Communicate" have been studied in the context of Decentralized-POMDP (Dec-POMDP) \cite{Roth2006,Carlin2009ValueOC} and Multi-agent RL (MARL) \cite{Zhu2022,Mao2020}. The goal in these studies is to simultaneously learn both a communication policy and an acting policy to maximize performance while adhering to communication constraints. The ``MARL with communications" framework is expected to play a vital role for 6G use cases involving multiple agents, massive sensory data, and constrained communication resources, especially when the learned communication policies lead to the emergence of agent-to-agent languages \cite{chafii2023emergent}. However, VoI is rarely used in this line of research \cite{Carlin2009ValueOC}. One negative consequence of this neglect is that it is challenging to explicitly express the trade-off between maximizing the performance of considered tasks and that of traditional communication services. As a result, communication constraints in these studies are often simplified, usually by limiting the size or number of messages, or by restricting which agents are allowed to transmit.\par

As wireless systems enter the AI era, the semantic communication paradigm, which emphasizes conveying the intended meaning of information rather than individual symbols, has emerged as a promising solution for enhancing wireless performance \cite{luo2022semantic}. Semantic communication and VoI-oriented communication share a common goal-oriented philosophy in communication system design. However, they differ in focus: semantic communication emphasizes understanding the meaning of transmitted content such as text, images, and speech, whereas VoI-oriented communication prioritizes the impact of information on the performance of control tasks supported by communication. Semantic communication is well-suited for non-control services, such as image and text sharing, but may fall short for control-related applications. For instance, in AD, semantic communication focuses on identifying objects like vehicles and roadblocks, while VoI-oriented communication goes further by determining which information is critical for the performance and safety of driving tasks. Therefore, integrating VoI-oriented communication with semantic communication offers a promising direction for supporting control-related services. \par

\subsubsection{Contribution 3}  
We propose a systematic framework to leverage VoI for optimizing communication systems. We classify communication decisions into three broad categories and identify the three time scales for decision-making. A general optimization objective is formulated to make VoI-oriented communication decisions, aiming to maximize communication performance for traditional services while minimizing performance degradation in vehicle control due to non-ideal communications. This performance degradation is quantified by the EVoI. \par 

In our framework, SSDP models are developed for the ``When", ``What", and ``How" to Communicate problems, with IVoI-associated reward functions derived from the optimization objective. Additionally, we detail the application of various VoI metrics in guiding different  types of communication decisions, providing a structured approach to optimize system-level performance. \par

The reminder of this paper is organized as follows. Section II describes the system model. Section III provides a brief introduction to the preliminaries of MDP. Section IV presents the SSDP model as a theoretical foundation, addressing Q1. Subsequently, Sections V and VI address Q2 by introducing the VoI definition and the DRL-based estimation method, respectively. Section VII addresses Q3 by presenting a VoI-oriented optimization framework for communication systems. Finally, Section VIII concludes the paper with a discussion on future research directions.\par

\section{System Model}
\begin{figure*}[!t]
	\centering
\includegraphics[width=0.9\textwidth]{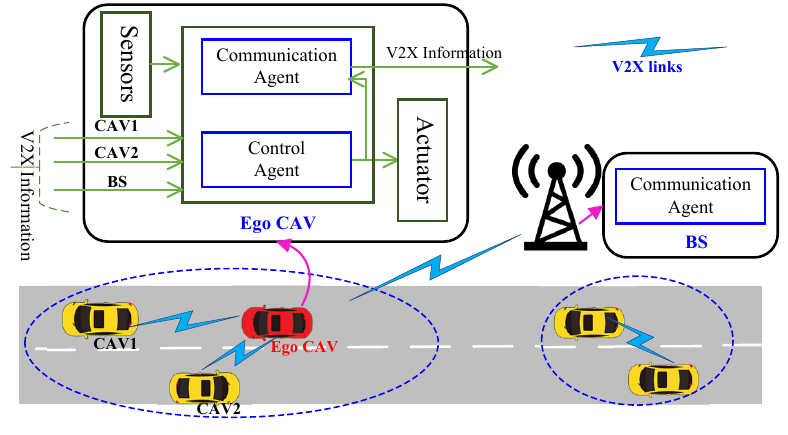}
\caption{Illustration of system model.}
\label{fig_system_model}
\end{figure*}

In this paper, we consider discrete time system where the time is discretized into equal-length control intervals indexed by $k \in \mathcal{K} = \{0,1,\cdots, \infty\}$. Without loss of generality, we focus on the ego CAV shown in Fig.~\ref{fig_system_model}. The ego CAV is equipped with a transmitter and receiver pair, multiple local on-board sensors (such as cameras, LiDARs and radars) and actuators, as well as a pair of control and communication agents. There are numerous remote sensors deployed on other vehicles and within the infrastructure. These remote sensors can transmit their collected data to the ego CAV using V2X communication, either directly via Vehicle-to-Vehicle (V2V) links or through a Base Station (BS) using Vehicle-to-Infrastructure (V2I) links. A communication agent operates at the BS, and potentially at some of the remote sensors.  \par

At each control interval $k$, the control agent determines the CAV control input $u_{k}$\footnote{We use $x_{k}$ to represent any variable $x$ at the beginning of control interval $k$. } based on the available sensory and V2X information. The control input $u_{k}$ is applied to the actuators at the beginning of control interval $k$ and holds constant within the time period of the control interval. \par

\begin{figure*} 
	\centering
    	\includegraphics[width=0.9\textwidth]{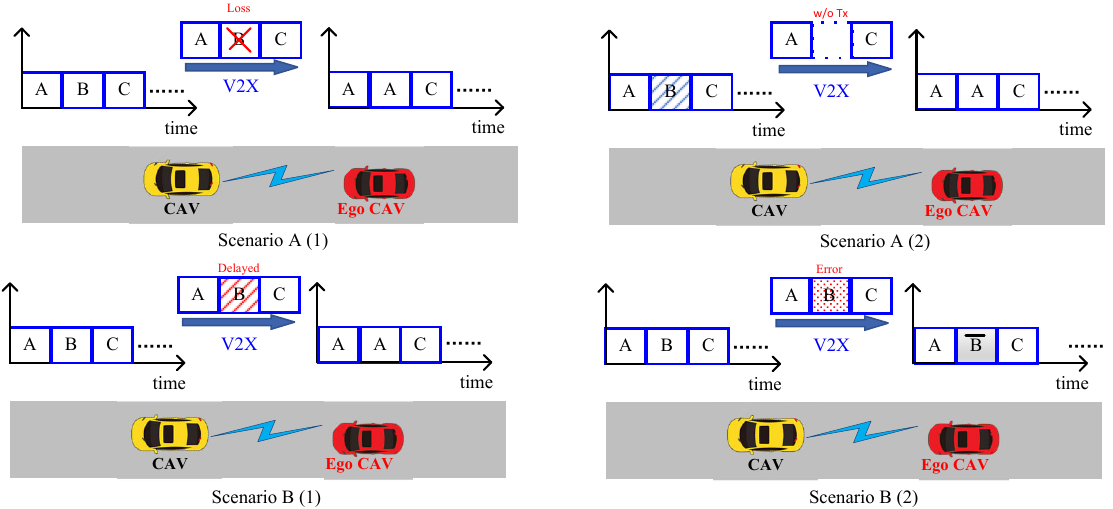}
    	\caption{Examples of typical scenarios.}
    	\label{fig_typicalscenarios}
\end{figure*}

Due to communication limitations in practical systems, the V2X information received by the control agent may be missing or imperfect. For illustration purposes, Fig. \ref{fig_typicalscenarios} presents examples of typical scenarios, which are explained as follows.\par

\textbf{Scenario A: Missing Information} \par
At control interval $k$, a piece of V2X information is not received by the ego CAV. The missing information can be caused by 
\begin{enumerate}
	\item the remote sensors withholding the information without transmission to save the usage of radio resources, or
	\item transmission loss in the communication systems.
\end{enumerate} 

In this case, the ego CAV could either
\begin{itemize}
	\item replace the missing information with some dummy bits; or
	\item replace the missing information with the last received information.
\end{itemize} 
The latter case converts the missing information to delayed information as described below.

\textbf{Scenario B: Imperfect Information} \par
The V2X information received by the ego CAV could be imperfect. Two main categories of imperfect information are listed below. \par

\begin{enumerate}

\item{Delayed Information:} 
At control interval $k$, the ego CAV receives delayed information due to transmission latency in the communication systems, or in cases where information is missing, it is replaced with the last received data, as previously described. This means that the information used by the control agent to make decisions at control interval $k$ was sampled at control interval $k-\tau_{k}$ by the remote sensors. $\tau_{k}$ is normally a random variable whose distribution depends on the stochastic properties of the wireless channel and RRM mechanisms.\par 

\item{Imprecise Information:}
At control interval $k$, a noisy version of the information is received by the CAV. Specifically, the received information may deviate from the transmitted information due to
\begin{itemize}
\item the remote sensors processing the information before transmission and thus results in information distortion, or
\item transmission errors in the communication systems.
\end{itemize} 

\end{enumerate}

Given the vast amount of remote sensory information and the impracticality of sharing all of it perfectly via V2X communication, this paper explores the answers to the three questions identified in Section I. \par

\newtheorem{case}{Case}
\begin{case}
	We consider a simple vehicle-following control problem with two vehicles, wherein the position, velocity and acceleration of a following vehicle (follower) $i$ are denoted by $p_{i}$, $v_{i}$, $acc_{i}$, respectively. Here $p_{i}$ represents the one-dimensional position of the center of the front bumper of vehicle $i$. \par

  The vehicle $i$ obeys the dynamics model described by a first-order system:
\begin{align}
\label{eq2}
& \dot{p}_{i}=v_{i},  \IEEEnonumber \\   
& \dot{v}_{i}=acc_{i},  \IEEEnonumber \\ 
& \dot{acc}_{i}=-\frac{1}{\rho_{i}}acc_{i}+\frac{1}{\rho_{i}}u_{i},
\end{align}

	\noindent where $\rho_{i}$ is a time constant representing the driveline dynamics of vehicle $i$ and $u_{i}$ is the vehicle's control input.
	
	We denote the headway of vehicle $i$, i.e., bumper-to-bumper distance between $i$ and its preceding vehicle (predecessor) $i-1$, by $d_{i}$, which satisfies
	\begin{equation}
	\label{eq5}
	d_{i}=p_{i-1}-p_{i}-L_{i-1},
	\end{equation}
	\noindent where $L_{i-1}$ is the the length of vehicle $i-1$.
	
	According to Constant Time-headway Policy (CTHP), vehicle $i$ aims for maintaining a desired headway of $d_{\sigma,i}$, given by
	\begin{equation}
	\label{eq6}
	d_{\sigma,i}=\sigma_{i}+h_{i}v_{i},
	\end{equation}
	\noindent where $\sigma_{i}$ is a constant standstill distance for vehicle $i$ and $h_{i}$ is the desired time-gap of vehicle $i$.
	
	The control errors, including position error $e_{pi}$ and velocity error $e_{vi}$ are defined as
	\begin{align}
	\label{eq7} 
&	e_{pi}=d_{i}-d_{\sigma,i}, \IEEEnonumber \\
&	e_{vi}=v_{i-1}-v_{i}.
	\end{align}

	The time horizon is discretized into $K$ control
intervals, each with a duration of $T$ seconds. Let $x_{i,k}=[e_{pi,k},e_{vi,k},acc_{i,k}]^{\mathrm{T}}$, where the subscript $k\in\{0,1,\cdots,K-1\}$ indicates the values of the variables at the beginning of control interval $k$. The system dynamics in discrete time are derived from \eqref{eq2} on the basis of forward Euler discretization:
	\begin{equation}
	\label{eq120}
	x_{i,k+1}=A_{i}x_{i,k}+B_{i}u_{i,k}+C_{i}acc_{i-1,k},
	\end{equation}
	\noindent where
	\begin{equation}
	\label{eq130}
	A_{i}=\begin{bmatrix}
	1 & T & -h_{i}T \\
	0 & 1 & -T \\
	0 & 0 & 1-\frac{T}{\rho_{i}}
	\end{bmatrix},
	B_{i}=\begin{bmatrix}
	0\\
	0 \\
	\frac{T}{\rho_{i}}
	\end{bmatrix},
	C_{i}=\begin{bmatrix}
	0\\
	T \\
	0
	\end{bmatrix}.		
	\end{equation}	
	
	At the beginning of each control interval, the control agent of vehicle $i$ has to determine the control input $u_{i,k}$ based on the observation of the state. The velocity $v_{i,k}$ and acceleration $acc_{i,k}$ can be measured locally, while the control errors $e_{pi,k}$ and $e_{vi,k}$ can be quantified by a radar unit mounted at the front of the vehicle. On the other hand, vehicle $i$ can only obtain the acceleration $acc_{i-1,k}$ of the predecessor through V2X communications. To determine the optimal action $u_{i,k}$, we must decide when and how to share information by V2X communications from predecessor $i-1$ to its follower $i$.

\end{case}

\section{Preliminaries - Markov Decision Process}
Since our focus is on AV controllers developed through MDP solutions, we provide an overview of MDP fundamentals to facilitate understanding of the subsequent contents in this paper.

   	An MDP is defined by the tuple $(\mathcal{S},\mathcal{A},P,r,\gamma)$, where $S_{k}\in\mathcal{S}$ and $a_{k}\in\mathcal{A}$ are state and action\footnote{The CAV control input $u_{k}$ is the action $a_{k}$ in the AV control problem.} at time step $k$ within state space $\mathcal{S}$ and action space $\mathcal{A}$, respectively. At each time step $k$, the action $a_{k}$ is determined by the agent based on the state $S_{k}$ and applied to the environment. Partly as a result of the $a_{k}$, a numerical reward signal $R_{k+1}\in\mathcal{R}$ is generated and the state $S_{k}$ transits to a new state $S_{k+1}\in\mathcal{S}$ at time step $k+1$. Both the next state and the reward are related to the current state and action, where a well-defined discrete probability distribution can be given as $\mathrm{Pr}(S_{k+1}, R_{k+1} | S_k, a_k)$ when the set of states, actions, and rewards all have a finite number of elements. The four-argument probability distribution defines the dynamics of the MDP, based on which the other functions of the environment can be derived, such as the state transition probability
   	\begin{equation}
   	\label{trans_prob}
   \mathrm{Pr}(S_{k+1}|S_{k},a_{k})=\sum_{r\in\mathcal{R}}\mathrm{Pr}(S_{k+1}, R_{k+1}=r | S_k, a_k),
   	\end{equation}
   	\noindent and the two-argument reward function
     \begin{align}	
   	\label{reward}
   	 r(S_{k},a_{k})&=\mathrm{E}[R_{k+1}|S_{k},a_{k}] \IEEEnonumber \\&=\sum_{r\in\mathcal{R}}r\sum_{s'\in\mathcal{S}}\mathrm{Pr}(S_{k+1}=s', R_{k+1}=r | S_k, a_k).
   	\end{align}
   	The state transition matrix $P$ defines state transition probabilities from all
   	states $S_{k}\in\mathcal{S}$ to all successor states $S_{k+1}\in\mathcal{S}$. \par
   	
   	 The agent uses a policy $\pi$ to determine the action $a_{k}$, where the policy is a conditional probability distribution over the action space given a state
   	\begin{equation}
   	\pi(a|s)=\mathrm{Pr}(a_{k}=a|S_{k}=s), \forall s\in\mathcal{S}, a\in\mathcal{A}
   	\end{equation}

   	The goal of the agent is to learn a policy that maximizes the cumulative reward it receives in the long run. There are different settings for formulating the cumulative reward in the theories of dynamic programming and RL, such as episodic setting, discounted setting, and average-reward setting. In this paper, we use a unified notation that covers both the episodic and discounted settings for explanatory purposes \cite{Sutton1998}, while our core ideas can be extended to other settings as well. To formally define the optimization objective of the agent, the state-value or value function $v_{\pi}(s)$ is defined as the expected cumulative reward from starting in state $s$ and then following policy $\pi$
   	\begin{equation}
   	\label{value}
   	v_{\pi}(s)=\mathrm{E}_{\pi}\left[\sum_{l=0}^{\infty}\gamma^{l} R_{k+l+1}\arrowvert S_{k}=s\right], \forall s\in\mathcal{S},
   	\end{equation}
   	\noindent where $\gamma \in [0, 1]$ is the discount factor, and $\mathrm{E}_{\pi}[\cdot]$ denotes the expected value of a random variable when the agent follows policy $\pi$. 

   	Similarly, the action-value or Q function $q_{\pi}(s,a)$ is defined as the expected cumulative reward from starting in state $s$, taking action $a$, and then following policy $\pi$
   	\begin{equation}
   	\label{q}
   	q_{\pi}(s,a)=\mathrm{E}_{\pi}\left[\sum_{l=0}^{\infty}\gamma^{l} R_{k+l+1}\arrowvert S_{k}=s,a_{k}=a\right], \forall s\in\mathcal{S}, a\in\mathcal{A}
   	\end{equation}
   	
   	Therefore, the objective of the control agent can be defined as deriving the optimal policy $\pi^{*}$ that maximizes the value function for all the states
   	\begin{equation}
   	v_{\pi^{*}}(s)=\sup_{\pi}v_{\pi}(s), \forall s\in\mathcal{S}.
   	\end{equation}

   	To quantify how good a policy is, the performance $J_{\pi}$ of a policy $\pi$ is defined as
   	\begin{equation}
   	\label{performance}
   	J_{\pi}=\mathrm{E}_{p(S_{0})}[v_{\pi}(S_{0})]=\mathrm{E}_{\pi,p(S_{0})}\left[\sum_{k=0}^{\infty}\gamma^{k} R_{k+1}\arrowvert S_{0}\right],
   	\end{equation}
    \noindent where $p(S_{0})$ denotes the probability distribution of the initial state $S_{0}$. 
   	
    Based on the value function $v_{\pi}(s)$ and Q function $q_{\pi}(s,a)$, we can derive the advantage function as
    \begin{equation}
    \label{advantage}
    A_{\pi}(s,a)=q_{\pi}(s,a)-v_{\pi}(s),
    \end{equation}
    \noindent which quantifies whether action $a$ is better or worse than the policy's default behavior.

For AV control tasks, state $S_{k}$ can be constructed based on the available sensory information. However, some useful information for decision-making may be missing from or imperfect in the state. Therefore,  we aim to quantify the benefit of making decisions when a piece of missing or imperfect information becomes available or perfect in the state, and we seek answers to Questions 1 and 2 outlined in Section I. \par

	\section{Sequential Stochastic Decision Process}
 
The MDP model faces some limitations in addressing Q1 and Q2 in Section I. 
One key issue is that the MDP model does not explicitly represent the useful information missing from the state. To develop a rigorous and practical method for defining and evaluating the VoI, we first extend the MDP model to a more general model known as the SSDP model, based on which the Q1 is addressed.

\subsection{SSDP Basics}
\newtheorem{mydef}{Definition}
\begin{mydef}[Sequential Stochastic Decision Process]
	Define an SSDP by a tuple $\mathfrak{S}=\{\mathcal{S},\mathcal{A},\mathcal{W},f^{S},r,f^{W}, \gamma\}$, where $\mathcal{S}$, $\mathcal{A}$, and $\gamma$ are the state space, action space, and discount factor as defined for an MDP, while 
	\begin{itemize}
		\item $W_{k}\in\mathcal{W}$ represents the exogenous information within its outcome space $\mathcal{W}$ that is not available at time step $k$ before the decision $a_{k}$ is made. The exogenous information $W_{k}$ includes any information that meets at lease one of the following two conditions: (1) $W_{k}$ affects the transition of state $S_{k}$, and (2) $W_{k}$ affects the reward $R_{k+1}$;
		\item $f^{S}$ is the system's state transition function governing $S_{k+1}=f^{S}(S_{k},a_{k},W_{k})$;
		\item $R_{k+1}=r(S_{k},a_{k},W_{k})$ is the reward function;
		\item $f^{W}$ is the function governing the exogenous information $W_{k}=f^{W}(\tilde{W}_{k})$, where $\tilde{W}_{k}\in\tilde{\mathcal{W}}$ represents all the deterministic and/or random parameters that affect the value of $W_{k}$.   	
	\end{itemize}
\end{mydef}

\newtheorem{case2}[case]{Case}
\begin{case2}
	\newtheorem{problem}{Problem}
	In Case 1, the vehicle-following control problem operating without V2X communications can be formulated as an SSDP $\{S_{i,k},a_{i,k},W_{i,k},f^{S_{i}},f^{W_{i}},R\}$  with 
	\begin{itemize}
		\item state $S_{i,k}=x_{i,k}=[e_{pi,k},e_{vi,k},acc_{i,k}]^{\mathrm{T}}$;
		\item action $a_{i,k}=u_{i,k}$;
		\item exogenous information $W_{i,k}=acc_{(i-1),k}$;
		\item state transition function $f^{S_{i}}$ governing 
		\begin{align}
		\label{eq11}
		S_{i,k+1}=f^{S_{i}}(S_{i,k},a_{i,k},W_{i,k}), 
		\end{align}	
		\noindent which is given by \eqref{eq120} and \eqref{eq130};
		\item exogenous information function $f^{W_{i}}$ given by
		\begin{align}
		\label{eq12}
		W_{i,k}&=f^{W_{i}}\left(W_{i,(k-1)},u_{(i-1),(k-1)}\right)\IEEEnonumber \\&=(1-\frac{1}{\rho_{i-1}})W_{i,(k-1)}+\frac{1}{\rho_{i-1}}u_{(i-1),(k-1)},
		\end{align}
		\noindent which is derived from the predecessor $i-1$'s dynamic model $\dot{acc}_{i-1}=-\frac{1}{\rho_{i-1}}acc_{i-1}+\frac{1}{\rho_{i-1}}u_{i-1}$ based on forward Euler discretization;
		\item and the reward function $r(S_{i,k},a_{i,k})$ given by 
\begin{align}
		\label{eq13}
		& r(S_{i,k},a_{i,k})= \IEEEnonumber \\
        &-\{|\frac{e_{pi,k}}{\hat{e}_{p,\mathrm{max}}}|+a|\frac{e_{vi,k}}{\hat{e}_{v,\mathrm{max}}}|+b|\frac{a_{i,k}}{u_{\mathrm{max}}}|+c|\frac{j_{i,k}}{2acc_{\mathrm{max}}/T}|\},
		\end{align}
		\noindent where $j_{i,k}$ is the change rate in acceleration, known as the jerk:
        \begin {align}	
	j_{i,k}&=\frac {acc_{i,k+1}-acc_{i,k}}{T}=-\frac{1}{\tau_{i}}acc_{i,k}+\frac{1}{\tau_{i}}a_{i,k},
        \end{align}
       \noindent and $\hat{e}_{p,\mathrm{max}}$ and $\hat{e}_{v,\mathrm{max}}$ are the nominal maximum control errors such that  it is larger than most possible control errors. $a$, $b$ and $c$ are the positive weights and can be adjusted to determine the relative importance of minimizing each term.\par 
       The reward function reflects the optimization objective: minimizing position error $e_{pi,k}$ and velocity error $e_{vi,k}$ to achieve the vehicle following target, while penalizing control input $u_{i,k}$ to reduce fuel consumption and jerk for improved driving comfort.\par
       To improve the performance of training a DRL agent, a Huber loss function can be used. It applies the squared error instead of the absolute error when the absolute reward value is below a certain threshold \cite{9951132}. \par
	\end{itemize}
	
\end{case2}

The SSDP is an extended version of MDP, incorporating the cases with non-Markov states. The following theorem provides the condition when the SSDP reduces to an MDP. The proof of Theorem 1 is provided in Appendix A.\par

\newtheorem{mythm}{Theorem}
\begin{mythm}[Relationship between SSDP and MDP]
The SSDP reduces to an MDP if and only if $\tilde{W}_{k}\subseteq\{S_{k},a_{k},\zeta_{k}\}$ in Definition 1, where $\zeta_{k}$ is a set of random variables whose distribution does not depend on the past states, actions and exogenous information, i.e., $\{S_{k'},a_{k'},W_{k'}\}_{k'\leq k-1}$. \par 

\end{mythm}

\newtheorem{case3}[case]{Case}
\begin{case3}
For the SSDP in Case 2, we have $\tilde{W}_{i,k}=\{W_{i,(k-1)},u_{(i-1),(k-1)}\}$. Therefore, this SSDP is not an MDP according to Theorem 1, since $\tilde{W}_{i,k}$ depends on the past exogenous information $W_{i,k-1}$.   
\end{case3}  	

\subsection{Augmented-State SSDP}

Given the SSDP in Definition 1, we aim to identify the set of additional information that can improve decisions beyond what is already included in the state. We start by inspecting the exogenous information $W_{k}$ in Definition 1, and examine whether its availability before decision making leads to better decisions. \par 

\newtheorem{mydef2}[mydef]{Definition}
\begin{mydef2}[Augmented-state SSDP with $W_{k}$]
	Assume that the exogenous information $W_{k}$ in the SSDP given in Definition 1 is available before decision $a_{k}$ is made. Then we define an augmented-state SSDP by $\tilde{\mathfrak{S}}=\{\tilde{\mathcal{S}},\mathcal{A},\tilde{\mathcal{W}},f^{\tilde{S}},r,f^{\tilde{W}},\gamma\}$, where the augmented state $\tilde{S}_{k}=(S_{k},W_{k})$ is obtained by including the exogenous information $W_{k}$ in the enlarged state at time step $k$. The action $a_{k}$, the reward function $r(\tilde{S}_{k},a_{k})=r(S_{k},a_{k},W_{k})$, and the discount factor $\gamma$ are the same as those of the original SSDP. 

	The system's state transition function $f^{\tilde{S}}$ becomes  
	\begin{align}
	\label{eq10}
	\tilde{S}_{k+1}&=
	\begin{pmatrix}
	S_{k+1}\\ W_{k+1}
	\end{pmatrix} \IEEEnonumber \\	
	& =   
	\begin{pmatrix}
	f^{S}(S_{k},a_{k},W_{k})\\ 
	f^{W}(\tilde{W}_{k+1})
	\end{pmatrix} \IEEEnonumber \\	
	&=f^{\tilde{S}}(\tilde{S}_{k},a_{k},\tilde{W}_{k+1}\backslash{\{\tilde{S}_{k},a_{k}\}}).
	\end{align} 
	
	From \eqref{eq10}, the exogenous information becomes $\tilde{W}_{k+1}\backslash{\{\tilde{S}_{k},a_{k}\}}$, where $\tilde{W}_{k+1}\in\tilde{\mathcal{W}}$. Finally, the exogenous information function becomes $\tilde{W}_{k+1}\backslash{\{\tilde{S}_{k},a_{k}\}}=f^{\tilde{W}}(\tilde{\tilde{W}}_{k+1})$, where $\tilde{\tilde{W}}_{k+1}\in\tilde{\tilde{\mathcal{W}}}$ represents all the parameters that affect the value of $\tilde{W}_{k+1}\backslash{\{\tilde{S}_{k},a_{k}\}}$.    
\end{mydef2}

\newtheorem{case4}[case]{Case}
\begin{case4}
	The vehicle-following control problem relying on V2X communications where $acc_{(i-1),k}$ is transmitted from the preceding vehicle $i-1$ can be formulated as an augmented-state SSDP $\{\tilde{S}_{i,k},a_{i,k},\tilde{W}_{i,k},f^{\tilde{S}_{i}},f^{\tilde{W}_{i}},R\}$ with 
\begin{itemize}
	\item state $\tilde{S}_{i,k}=[e_{pi,k},e_{vi,k},acc_{i,k},acc_{(i-1),k}]^{\mathrm{T}}=[(S_{i,k})^{\mathrm{T}},W_{i,k}]^{\mathrm{T}}$;
	\item exogenous information $\tilde{W}_{i,k}=u_{(i-1),k}$;
	\item state transition function $f^{\tilde{S}_{i}}$ given by
	\begin{align}
	\label{eq27}
	\tilde{S}_{i,k+1} &= \begin{pmatrix}
	S_{i,k+1}\\ W_{i,k+1}
	\end{pmatrix}
	=
	\begin{pmatrix}
	f^{S_{i}}(S_{i,k},a_{i,k},W_{i,k})\\ 
	f^{W_{i}}(W_{i,k},u_{(i-1),k}),
	\end{pmatrix}  \IEEEnonumber \\
	&=f^{\tilde{S}_{i}}(\tilde{S}_{i,k},a_{i,k},u_{(i-1),k})
	\end{align} 
	\noindent where $f^{S_{i}}$ is formulated in \eqref{eq11}, and $f^{W_{i}}$ is given in \eqref{eq12};
	\item exogenous information function $f^{\tilde{W}_{i}}$ given by
	\begin{equation}
	\label{eq28}
	u_{(i-1),k}=f^{\tilde{W}_{i}}(\tilde{\tilde{W}}_{i,k}).
	\end{equation}
 \noindent where $\tilde{\tilde{W}}_{i,k}$ is the set of information that affects the control input $u_{(i-1),k}$ of predecessor $i-1$, such as the road conditions and the controller parameters of predecessor $i-1$.
\end{itemize}
\end{case4} 

\newtheorem{mythm2}[mythm]{Theorem}
\begin{mythm2}
	The optimal policy of the augmented-state SSDP $\tilde{\pi}^{*}(\tilde{S}_{k})$ is at least as good as that of the original SSDP $\pi^{*}(S_{k})$. 
\end{mythm2}  

The proof of Theorem 2 is provided in Appendix B. Theorem 2 demonstrates that including exogenous information $W_{k}$ in the augmented state $\tilde{S}_{k}$ has the potential to improve the decisions. Alternatively, in cases where $W_{k}$ is unavailable, the knowledge of $\tilde{W}_{k}$ can be leveraged to conceive an augmented-state SSDP, since $W_{k}=f^{W}(\tilde{W}_{k})$.

\newtheorem{mydef3}[mydef]{Definition}
\begin{mydef3}[Augmented-state SSDP with $\tilde{W}_{k}$]
	Assume that $\tilde{W}_{k}$ instead of $W_{k}$ in the SSDP given in Definition 1 is available before decision $a_{k}$ is made. Then we define an augmented-state SSDP by $\tilde{\mathfrak{S}}'=\{\tilde{\mathcal{S}}',\mathcal{A},\tilde{\tilde{\mathcal{W}}},f^{\tilde{S}'},r,f^{\tilde{\tilde{W}}},\gamma\}$, where the augmented state $\tilde{S}'_{k}=(S_{k},\tilde{W}_{k})$ is obtained by including $\tilde{W}_{k}$ in the enlarged state at time step $k$. The action $a_{k}$, the reward function $r(\tilde{S}_{k},a_{k})=r(S_{k},a_{k},f^{W}(\tilde{W}_{k}))$, and the discount factor $\gamma$ are the same as those of the original SSDP. 

	The system's state transition function $f^{\tilde{S}}$ becomes  
	\begin{align}
	\label{eq20}
	\tilde{S}_{k+1}&=
	\begin{pmatrix}
	S_{k+1}\\ \tilde{W}_{k+1}
	\end{pmatrix} \IEEEnonumber \\	
	& =   
	\begin{pmatrix}
	f^{S}(S_{k},a_{k},f^{W}(\tilde{W}_{k}))\\ 
	f^{\tilde{W}}(\tilde{\tilde{W}}_{k+1},S_{k},a_{k},f^{W}(\tilde{W}_{k}))
	\end{pmatrix} \IEEEnonumber \\	
	&=f^{\tilde{S}'}(\tilde{S}'_{k},a_{k},\tilde{\tilde{W}}_{k+1}\backslash\{\tilde{S}'_{k},a_{k}\}).
	\end{align} 
	
	From \eqref{eq20}, the exogenous information becomes $\tilde{\tilde{W}}_{k+1}\backslash\{\tilde{S}'_{k},a_{k}\}$, where $\tilde{\tilde{W}}_{k+1}\in\tilde{\tilde{\mathcal{W}}}$. Finally, the exogenous information function becomes $\tilde{\tilde{W}}_{k+1}\backslash\{\tilde{S}'_{k},a_{k}\}=f^{\tilde{\tilde{W}}}(\tilde{\tilde{\tilde{W}}}_{k+1})$, where $\tilde{\tilde{\tilde{W}}}_{k+1}\in\tilde{\tilde{\tilde{\mathcal{W}}}}$ represents all the parameters that affect the value of $\tilde{\tilde{W}}_{k+1}\backslash\{\tilde{S}'_{k},a_{k}\}$.    
\end{mydef3}

The following Lemma 1 identifies the set of additional information that can improve decisions.\par

\newtheorem{mylem}{Lemma}
\begin{mylem}
Given the SSDP in Definition 1, the set of information $\mathcal{I}_k\in\{W_{k},\tilde{W}_{k},\tilde{\tilde{W}}_{k},\cdots\}$ is beneficial for enhanced decision-making if becomes available before decision $a_k$ is made. Here, $W_{k}$ represents the exogenous information, $\tilde{W}_{k}$ represents the predictive information of $W_{k}$, and $\tilde{\tilde{W}}_{k}$ is the predictive information of $\tilde{W}_{k}$, and so on. Specifically, the availability of:
\begin{itemize}
    \item Exogenous information $W_{k}$,
    \item Predicative information $\tilde{W}_{k}$ when $W_{k}$ is unavailable,
    \item Predicative information $\tilde{\tilde{W}}_{k}$ when both $W_{k}$ and $\tilde{W}_{k}$ are unavailable, and so on
\end{itemize}
can improve decisions. Any information $I_k$ cannot enhance decision-making if $I_k\notin\mathcal{I}_k$, for all $\mathcal{I}_k\in\{W_{k},\tilde{W}_{k},\tilde{\tilde{W}}_{k},\cdots\}$.  
\end{mylem}

\newtheorem{remark}{Remark}
\begin{remark}[how to conceive the state $S_k$ in Definition 1]
$S_k$ is constructed based on the information available to the agent. However, not all information is necessarily relevant to decision-making. Lemma 1 provides a method to formulate the state in any MDP or SSDP model:
\begin{itemize}
    \item Step 1: Pick any available information that is necessary and sufficient to compute the reward $R_{k+1}$ and form $S_k$.
    \item Step 2: For each piece of available information $I_k$:
    \begin{itemize}
    \item if $I_k$ is necessary and sufficient to compute the next state $S_{k+1}$, update $S_k\leftarrow\{S_k,I_k\}$.
    \end{itemize}
\end{itemize}
\end{remark}

\newtheorem{remark2}[remark]{Remark}
\begin{remark2}[Relationship between Lemma 1 and the environment state]
The environment state introduced by Silver \cite{silver} is essentially the augmented state $\tilde{S}_{k}=(S_{k},W_{k})$ as defined in Definition 2, which incorporates the state $S_{k}$ and exogenous information $W_{k}$. Once the augmented state $\tilde{S}_{k}$ is known, the next state $S_{k+1}$ and reward $R_{k+1}$ are completely determined. All the information in the environment state is useful in making decisions, although some information may be more important than others. We will quantify the VoI in the environment state in Section IV. Given the environment state, no additional information can enhance decision-making. Therefore, the environment state includes the necessary and sufficient information for making the optimal decisions.
\end{remark2}

\newtheorem{remark3}[remark]{Remark}
\begin{remark3}[Relationship between Lemma 1 and the state definition by Powell]
According to Lemma 1, Powell's state definition can be revised to ``A state variable is the minimally dimensioned function of history that is necessary and sufficient to compute the transition function and the reward function." The decision or policy function is removed from the definition. Actually, if a piece of information is necessary and sufficient to compute the transition function and the reward function, it is useful to compute the policy function.     
\end{remark3}

Notice that the augmented-state SSDPs in Definition 2 and Definition 3 consider that the full set of information $W_k$ or $\tilde{W}_k$ is available before decision making. In general, we can conceive an augmented-state SSDP to incorporate any information $I_k\in\mathcal{I}_k$ in the augmented state $\tilde{S}_k$. \par

\subsection{Augmented-State SSDP with imperfect information}
Given the SSDP in Definition 1, we consider a more complex scenario where a piece of information $I_k\in\mathcal{I}_k$ is available before decision making but could be imperfect. Let $I_k^{\rm ip}$ denote the imperfect information. As discussed in Section II, the imperfect information can be divided into 
(1) delayed information, i.e., $I_k^{\rm ip}=I_{k-\tau_{k}}$; and (2) imprecise information, i.e., $I_k^{\rm ip}=I_k+Z_{k}$, where $Z_{k}$ is a random disturbance.

In either case, we can conceive augmented-state SSDPs to leverage the imperfect information for making improved decisions. In the following, as an example, we formulate an augmented-state SSDP where the exogenous information $W_{k}$ is delayed. 

\newtheorem{mydef4}[mydef]{Definition}
\begin{mydef4}[Random Delay SSDP]
	Assume that the delayed exogenous information $W_{k-\tau_{k}}$ in the SSDP given in Definition 1 is available before decision $a_{k}$ is made. Then we define a random delay SSDP by $\tilde{\mathfrak{S}}^{\rm d}=\{\tilde{\mathcal{S}}^{\rm d},\mathcal{A},\tilde{\mathcal{W}}^{\rm d},f^{\tilde{S}^{\rm d}},r,f^{\tilde{W}^{\rm d}},\gamma\}$, where the augmented state $\tilde{S}_{k}^{\rm d}=(S_{k-\tau_{k}},W_{k-\tau_{k}},\{a_{k'}\}_{k'=k-\tau_{\mathrm{max}}}^{k-1},\tau_{k})$ is obtained by including delayed exogenous information $W_{k-\tau_{k}}$, the last $\tau_{\mathrm{max}}$ actions $\{a_{k'}\}_{k'=k-\tau_{\mathrm{max}}}^{k-1}$, as well as the observation delay $\tau_{k}$ in the enlarged state at time step $k$. The action $a_{k}$, the reward function $r(S_{k},a_{k},W_{k})$, and the discount factor $\gamma$ are the same as those of the original SSDP. 

	The system's state transition function $f^{\tilde{S}}$ becomes

\newcounter{mytempeqncnt}
\begin{figure*}[hb]
\normalsize
\hrulefill
\setcounter{mytempeqncnt}{\value{equation}}
\setcounter{equation}{24}
\begin{align}
\label{eq17}
	\tilde{S}_{k+1}^{\rm d}&=
	\begin{pmatrix}
	\tilde{S}_{k+1-\tau_{k+1}} \\ \{a_{k'}\}_{k'=k+1-\tau_{\mathrm{max}}}^{k} \\ \tau_{k+1}
	\end{pmatrix} \IEEEnonumber \\	
	& =   
	\begin{pmatrix}
	f^{S}(\cdots f^{S}(f^{S}(\tilde{S}_{k-\tau_{k}},a_{k-\tau_{k}},\tilde{W}_{k-\tau_{k}}),a_{k-\tau_{k}+1},\tilde{W}_{k-\tau_{k}+1})\cdots,a_{k-\tau_{k+1}},\tilde{W}_{k-\tau_{k+1}})\\ 
	\{a_{k'}\}_{k'=k+1-\tau_{\mathrm{max}}}^{k-1},a_{k} \\
	f^{\tau}(\tau_{k},W^{\tau}_{k})
	\end{pmatrix} \IEEEnonumber \\	
	&=f^{\tilde{S}}(\tilde{S}_{k}^{\rm d},a_{k},\tilde{W}_{k}^{\rm d}).
\end{align}
\setcounter{equation}{\value{mytempeqncnt}+1}
\end{figure*}

	From \eqref{eq17}, the exogenous information becomes $\tilde{W}_{k}^{\rm d}=(\{\tilde{W}_{k'}\}_{k'=k-\tau_{k}}^{k-\tau_{k+1}},W^{\tau}_{k})\in\tilde{\mathcal{W}}^{\rm d}$. Finally, the exogenous information function is $\tilde{W}_{k}^{\rm d}=f^{\tilde{W}^{\rm d}}(\tilde{\tilde{W}}_{k}^{\rm d})$, where $\tilde{\tilde{W}}_{k}^{\rm d}\in\tilde{\tilde{\mathcal{W}}}^{\rm d}$ represents all the parameters that affect the value of $\tilde{W}_{k}^{\rm d}$.    
\end{mydef4}

\newtheorem{case5}[case]{Case}
\begin{case5}
A Random Delay SSDP can be formulated for with state

$\tilde{S}_{i,k}^{\rm d}=[x_{i,k-\tau_{i,k}},acc_{(i-1),k-\tau_{i,k}},\{u_{i,k'}\}_{k'=k-\tau_{\mathrm{max}}}^{k-1},\tau_{i,k}]^{\mathrm{T}}$.	
\end{case5} 

In the rest of the paper, we use $\tilde{S}^{\rm ip}$ to denote an augmented state with imperfect information. $\tilde{S}^{\rm d}_{k}$ defined in Definition 4 is an example of $\tilde{S}^{\rm ip}$. \par

\section{VoI Definition}

In Section IV, we have identified the additional set of information beyond the state variables that empowers the control agent to enhance decision-making. In this section, we formally define the VoI to quantify the merit of information. Broadly speaking, two categories of VoI are introduced: utility-based VoI and information theory-based VoI. \par

\subsection{Utility-based VoI}
\subsubsection{Value of Missing Information vs. Value of Imperfect Information}

Since the ultimate objective for a control agent in obtaining information is to make improved decisions for achieving better control performance, the utility-based VoI measures the performance difference when decisions are made based on partial/imperfect versus full/perfect information. \par

Recall the state $S_{k}$ includes the set of information that is currently available to the agent. Consider the case when an additional piece of information $I_{k}\in\mathcal{I}_k$ becomes available before decision-making. Two scenarios are considered: (1) $I_{k}$ is perfectly available; (2) imperfect information $I_{k}^{\rm ip}$ is available instead of $I_{k}$. \par

We define two types of VoI, namely the \emph{Value of Missing Information} (VoMI) and the \emph{Value of Imperfect Information} (VoII), to measure the value of $I_{k}$ in these two scenarios, respectively. \par

\newtheorem{mydef5}[mydef]{Definition}
\begin{mydef5}{(Value of Missing Information (VoMI))}
	The VoMI is defined as the difference in performance between the acting policy $\pi$ for the SSDP, where the state $S_{k}$ excludes information $I_{k}$, and the optimal policy $\tilde{\pi}^{*}$ for the augmented-state SSDP, where the augmented state $\tilde{S}_{k}=\{S_{k},I_{k}\}$ includes $I_{k}$. 
\end{mydef5}

\newtheorem{mydef6}[mydef]{Definition}
\begin{mydef6}{(Value of Imperfect Information (VoII))}
	The VoII is defined as the difference in performance between the acting policy $\tilde{\pi}^{\rm ip}$ for the augmented-state SSDP with imperfect information, where the state $\tilde{S}_{k}^{\mathrm{ip}}$ is constructed from imperfect information $I_{k}^{\rm ip}$, and the optimal policy $\tilde{\pi}^{*}$ for the augmented-state SSDP, where the augmented state $\tilde{S}_{k}=\{S_{k},I_{k}\}$ includes perfect information $I_{k}$. 	
\end{mydef6}

Note that both VoMI and VoII quantify the performance gap between a policy based on partial/imperfect observation (state with partial/imperfect information) and a policy based on full/perfect observation (state with full/perfect information). For ease of discussion, we introduce unified notations $\pi^{\rm{inf}}$ and $\pi^{\rm{sup}}$ to denote the policies in the VoMI and VoII definitions based on partial/imperfect and full/perfect observations, respectively - where $\pi^{\rm{inf}}$ yields inferior performance and $\pi^{\rm{sup}}$ exhibits superior performance. Correspondingly, the states are denoted as $S_{k}^{\rm{inf}}$ for partial/imperfect observations and $S_{k}^{\rm{sup}}$ for full/perfect observations.  \par

Taking VoMI as an example, $\pi^{\rm{inf}}$ could be the optimal policy $\pi^{*}$ for the SSDP with the state $S_{k}$ that excludes $I_{k}$. Alternatively, it could be derived from the optimal policy $\tilde{\pi}^{*}$ for the augmented-state SSDP, where $I_{k}$ in $\tilde{S}_{k} = \{S_{k}, I_{k}\}$ is replaced with dummy bits. More generally, it may represent any policy employed by the control agent to make decisions based on the partial observation $S_{k}$.\par


\newtheorem{remark4}[remark]{Remark}
\begin{remark4}{(Dependence of VoI on policies)}
In Definitions 5 and 6, we assume that the control agent follows policy $\pi^{\rm{inf}}$ when only partial or imperfect observations are available, and switches to policy $\pi^{\rm{sup}}$ when full or perfect observations are accessible. In other words, the control agent adopts a mixed policy, utilizing $\pi^{\rm{inf}}$ and $\pi^{\rm{sup}}$ depending on the quality of the available information, and is assumed to fully exploit the full/perfect information to make more informed decisions. In this setting, the dependence of VoI on the policies is solely determined by the robustness of $\pi^{\rm{inf}}$ to missing or imperfect information: the more robust $\pi^{\rm{inf}}$ is, the less valuable the additional or perfect information becomes. \par

When $\pi^{\rm{sup}}$ is not available to the control agent, $\pi^{\rm{inf}}$ is used in place of $\pi^{\rm{sup}}$ in both Definitions 5 and 6. In this case, the control agent always employs $\pi^{\rm{inf}}$, regardless of whether it receives partial/imperfect or full/perfect observations. Here, the VoI also depends on how effectively the agent can leverage the additional or perfect information using $\pi^{\rm{inf}}$: the less capable the agent is at utilizing this information, the less valuable it becomes to the agent, even though it may be more useful to another policy that can exploit it more effectively.
\end{remark4}

\subsubsection{Expected Cumulative VoI vs. Immediate VoI}
Depending on how the performance difference is measured, both VoMI and VoII can be further divided into two categories: Expected Cumulative VoI (EVoI) and Immediate VoI (IVoI). \par	

\newtheorem{mydef7}[mydef]{Definition}
\begin{mydef7}{(Expected Cumulative VoI (EVoI))}
The EVoI $\bar{\Xi}$ is defined as the difference in performance between the acting policy $\pi^{\rm{inf}}$ based on partial/imperfect observations and the optimal policy $\pi^{\rm{sup}}$ based on full/perfect observations, respectively. Formally,
\begin{equation}
\label{EVoI}
\bar{\Xi}=J_{\pi^{\rm{inf}}}-J_{\pi^{\rm{sup}}},
\end{equation}
\noindent where both $J_{\pi^{\rm{inf}}}$ and $J_{\pi^{\rm{sup}}}$ are calculated according to \eqref{performance}.
\end{mydef7}

The EVoI quantifies the performance loss as a consequence of the control agent consistently receiving the partial/imperfect observations at every control interval. In the terminology of RL, EVoI corresponds to the regret of $\pi^{\rm{inf}}$. It is worth noting that the EVoI is a non-positive value, with a larger EVoI indicating less performance loss. When the EVoI reaches its maximum value of zero, no performance loss occurs, implying that the missing or imperfect information has no impact on decision-making.\par

\newtheorem{mydef8}[mydef]{Definition}
\begin{mydef8}{(Immediate Value of Information (IVoI))}
Given a full/perfect observation $S_{k}^{\rm{sup}}$ and a partial/imperfect observation $S_{k}^{\rm{inf}}$ at control interval $k$, the IVoI $\xi_k$ is defined as the performance difference resulting from executing two different actions at interval $k$: one chosen by the acting policy $\pi^{\rm{inf}}$ based on $S_{k}^{\rm{inf}}$ and the other by the optimal policy $\pi^{\rm{sup}}$ based on $S_{k}^{\rm{sup}}$, while following $\pi^{\rm{sup}}$ based on full/perfect observations $S_{k+1}^{\rm{sup}},S_{k+2}^{\rm{sup}},\dots$ thereafter. Formally, 
	\begin{align}
	\label{OVoI}
\xi_k=A_{\pi^{\rm{sup}} }(S_{k}^{\rm{sup}}, a_{k}^{\rm{inf}})|_{a_{k}^{\rm{inf}}\sim\pi^{\rm{inf}}(\cdotp|S_{k}^{\rm{inf}})},
	\end{align}
\noindent where $A_{\pi^{\rm{sup}}}$ denotes the advantage function under the optimal policy $\pi^{\rm{sup}}$, and the action $a_{k}^{\rm{inf}}$ is sampled according to $\pi^{\rm{inf}}$ based on $S_{k}^{\rm{inf}}$.
\end{mydef8}

The IVoI indicates the performance loss as a consequence of the control agent receiving a partial/imperfect observation only at a specific control interval $k$, while consistently obtaining full/perfect  observations throughout the remaining time. It is defined for a given pair consisting of the full/perfect observation $S_{k}^{\rm{sup}}$ and inferior action $a_{k}^{\rm{inf}}$, which is prescribed by the acting policy $\pi^{\rm{inf}}$ based on the partial/imperfect observation $S_{k}^{\rm{inf}}$. In other words, IVoI measures the performance degradation when the agent, despite the full/perfect observation being $S_{k}^{\rm{sup}}=s$, relies on the partial/imperfect observation $S_{k}^{\rm{inf}}$ to select its action at control interval $k$. In the terminology of RL, IVoI corresponds to the advantage function of the optimal policy $\pi^{\rm{sup}}$. Similar to EVoI, IVoI is also a non-positive value, where maximizing IVoI leads to minimizing the performance loss.\par
	
It is proved in \cite{10376461} that the EVoI is equivalent to the expected cumulative IVoI, similar to the relationship between the expected cumulative reward and immediate reward in RL. 
\newtheorem{mylem2}[mylem]{Lemma}
\begin{mylem2}
	The EVoI equals to the expected cumulative IVoI, i.e.,
	\begin{align}
	\label{perf_diff}
	\bar{\Xi}=\mathrm{E}_{\pi^{\rm CM}}\mathrm{E}_{\pi^{\rm inf}}\left[\sum_{k=0}^{\infty} \gamma^{k}\xi_k\right],
	\ 0 \leq \gamma \leq 1,
	\end{align}
\end{mylem2}

\noindent where $\pi^{\rm CM}$ denotes the communication policy. Note that $\pi^{\rm CM}$ affects the stochastic properties of communication
nonideality, such as delay distribution, which in turn impacts
the transition of the partial/impaired observations received by the control agent. It is essential to clarify that the term \emph{Immediate} VoI doesn't imply its sole focus on quantifying the disparity in the \emph{immediate} rewards at control interval $k$, when the control decisions are made based on the partial/imperfect versus full/perfect observations, respectively. Since the decision at control interval $k$ not only affects the immediate reward at the current control interval but also the future rewards through its impact on the next state, the IVoI measures the performance loss in terms of the expected cumulative reward just as the EVoI does. \par

\begin{figure*}
	\centering
	\includegraphics[width=0.7\textwidth]{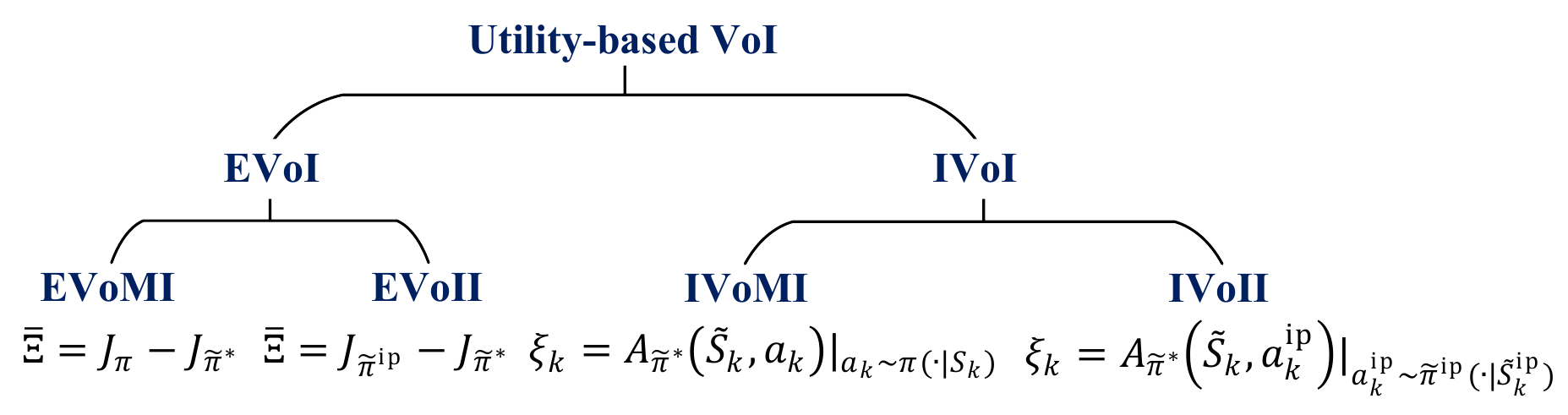}
	\caption{Four categories of Utility-based VoI.}
	\label{category}
\end{figure*}

Therefore, utility-based VoI can be categorized into two main types: EVoI and IVoI. Each of these types is further divided into two subcategories: VoMI and VoII. As illustrated in Fig. \ref{category}, this classification results in four distinct categories of utility-based VoI: Expected VoMI (EVoMI),  Expected VoII (EVoII), Immediate VoMI (IVoMI), and Immediate VoII (IVoII). According to Definition 5 and 6, the policy/state/action $\pi^{\rm{inf}}$/$S_k^{\rm{inf}}$/$a_k^{\rm{inf}}$ in \eqref{EVoI} and \eqref{OVoI} is replaced by $\pi$/$S_k$/$a_k$ and $\tilde{\pi}^{\rm ip}$/$\tilde{S}_k^{\rm ip}$/$a_k^{\rm ip}$ for VoMI and VoII, respectively, while $\pi^{\rm{sup}}$/$S_k^{\rm{sup}}$ in \eqref{EVoI} and \eqref{OVoI} is replaced by $\tilde{\pi}^{*}$/$\tilde{S}_k$.

%

\newtheorem{case6}[case]{Case}
\begin{case6}
Consider the vehicle-following control problem in Case 1. The control agent of vehicle $i$ can observe its position error $e_{pi,k}$, velocity error $e_{vi,k}$, and acceleration $acc_{i,k}$ based on local sensory information. Meanwhile, the acceleration $acc_{i-1,k}$ of the predecessor $i-1$ is remote sensory information that has to be transmitted from vehicle $i-1$ to vehicle $i$. In order to quantify the benefit of making decisions when $acc_{i-1,k}$ is available to the control agent of vehicle $i$, the four utility-based VoI can be defined for $acc_{i-1,k}$:
\begin{itemize}
 \item EVoMI $\Xi_{i}^{\rm M}=J_{\pi_{i}}-J_{\tilde{\pi}_{i}^{*}}$, where $J_{\pi_{i}}$ and $J_{\tilde{\pi}_{i}^{*}}$ are the performance of the acting policy $\pi_{i}$ based on partial observations $S_{i,k}=[e_{pi,k},e_{vi,k},acc_{i,k}]^{\mathrm{T}}$ (for the SSDP in Case 2) and the optimal policy $\tilde{\pi}_{i}^{*}$ based on full observations $\tilde{S}_{i,k}=[e_{pi,k},e_{vi,k},acc_{i,k},acc_{(i-1),k}]^{\mathrm{T}}$ (for the augmented-state SSDP in Case 4), respectively. The EVoMI quantifies the performance loss incurred when the following vehicle $i$ consistently fails to receive the acceleration $acc_{i-1,k}$ of the preceding vehicle $i-1$ at every control interval $k$ throughout the entire episode. Note that the performance refers to the expected cumulative reward defined in \eqref{performance}, while the reward function of the vehicle-following control problem, given in \eqref{eq13}, depends on the position error, velocity error, and acceleration of vehicle $i$.
 \item IVoMI $\xi_{i,k}^{\rm M}=A_{\tilde{\pi}_{i}^{*}}(\tilde{S}_{i,k},u_{i,k})|_{u_{i,k}\sim\pi_{i}(\cdot|S_{i,k})}$, where $A_{\tilde{\pi}_{i}^{*}}(\tilde{S}_{i,k},u_{i,k})$ is the advantage function of the optimal policy $\tilde{\pi}_{i}^{*}$, and $u_{i,k}\sim\pi_{i}(\cdot|S_{i,k})$ is the control input of vehicle $i$ that is sampled according to the acting policy $\pi_{i}(\cdot|S_{i,k})$ based on partial observation $S_{i,k}$. The IVoMI quantifies the performance loss incurred when the following vehicle $i$ fails to receive the acceleration $acc_{i-1,k}$ of the preceding vehicle $i-1$ at a specific control interval $k$, while consistently receiving this information during the rest of the episode. Consequently, the IVoMI is a function of the state action pair $\{\tilde{S}_{i,k},u_{i,k}\}$, where $\tilde{S}_{i,k}$ represents the full observation at control interval $k$, including $acc_{i-1,k}$. It measures the performance loss incurred when the control agent has access only to the partial observation (excluding $acc_{i-1,k}$) and selects action $u_{i,k}$ using the acting control policy. Unlike EVoI, IVoI distinguishes the value of knowing the predecessor's acceleration based on both its specific values and the available system state information, such as position error, velocity error, and acceleration of the follower. 
 \item EVoII $\Xi_{i}^{\rm I}=J_{\pi_{i}^{\rm d}}-J_{\tilde{\pi}_{i}^{*}}$, where $J_{\pi_{i}^{\rm d}}$ and $J_{\tilde{\pi}_{i}^{*}}$ are the performance of the acting policy $\pi_{i}^{\rm d}$ based on imperfect (delayed) observations $\tilde{S}_{i,k}^{\rm d}=[x_{i,k-\tau_{i,k}},acc_{(i-1),k-\tau_{i,k}},\{u_{i,k'}\}_{k'=k-\tau_{\mathrm{max}}}^{k-1},\tau_{i,k}]^{\mathrm{T}}$ (for the Random Delay SSDP in Case 5) and the optimal policy $\tilde{\pi}_{i}^{*}$ based on perfect observations $\tilde{S}_{i,k}=[e_{pi,k},e_{vi,k},acc_{i,k},acc_{(i-1),k}]^{\mathrm{T}}$ (for the augmented-state SSDP in Case 4), respectively. The EVoII is similar to the EVoMI, with the only difference being that the performance loss results from the following vehicle $i$ receiving a delayed acceleration $acc_{i-1,k-\tau_{k}}$ from the predecessor $i-1$, rather than not receiving it at all.
 \item IVoII $\xi_{i,k}^{\rm I}=A_{\tilde{\pi}_{i}^{*}}(\tilde{S}_{i,k},u_{i,k})|_{u_{i,k}\sim\tilde{\pi}_{i}^{\rm d}(\cdot|\tilde{S}_{i,k}^{\rm d})}$, where $A_{\tilde{\pi}_{i}^{*}}(\tilde{S}_{i,k},u_{i,k})$ is the advantage function of the optimal policy $\tilde{\pi}_{i}^{*}$, and $u_{i,k}\sim\tilde{\pi}_{i}^{\rm d}(\cdot|\tilde{S}_{i,k}^{\rm d})$ is the control input of vehicle $i$ that is sampled according to the acting policy $\tilde{\pi}_{i}^{\rm d}(\cdot|\tilde{S}_{i,k}^{\rm d})$ based on delayed observation $\tilde{S}_{i,k}^{\rm d}$. The IVoII is similar to the IVoMI, in the same way that the EVoII parallels the EVoMI.
\end{itemize}
\end{case6} 

\subsection{Information-Theory-based VoI}

According to Lemma 1, given the SSDP in Definition 1, a piece of information $I_k$ can improve decision making only if it is part of the exogenous information $W_{k}$ or helps in predicting $W_{k}$. Furthermore, as per Definition 1, $W_{k}$ must either affect the transition of state $S_{k}$ and/or the reward $R_{k+1}$. Thus, it follows intuitively that the value of $I_k$ is related to its impact on both the state transition and the reward. VoI can thus be defined by quantifying, ``\emph{How much better could we predict the next state $S_{k+1}$ or the reward $R_{k+1}$ if we included $I_{k}$ in the augmented-state $\tilde{S}_{k}$, compared to if we did not}?" \cite{lei2022deep}. \par

\newtheorem{mydef9}[mydef]{Definition}
\begin{mydef9}{(Information-Theory-based VoI (ITVoI))}
For the SSDP where the state $S_{k}$ does not include information $I_{k}$ and the augmented-state SSDP where the augmented state $\tilde{S}_{k}=\{S_{k},I_{k}\}$ includes $I_{k}$, we convert the state transition functions, the exogenous information function for $I_{k}$, and the reward function, i.e., $f^{S}$, $f^{\tilde{S}}$, $f^{I}$, and $r$ to the corresponding transition probabilities $T^{S}=p\{S_{k+1}|S_{k},a_{k}\}$, $T^{\tilde{S}}=p\{S_{k+1},I_{k+1}|S_{k},a_{k},I_{k}\}$, and $T^{I}=p\{I_{k+1}|I_{k},S_{k},a_{k}\}$, $T^{r(S)}=p\{R_{k+1}|S_{k},a_{k}\}$, and $T^{r(\tilde{S})}=p\{R_{k+1}|S_{k},I_{k},a_{k}\}$. The ITVoI is defined as the sum of the conditional KL divergence of $T^{S}\otimes T^{I}$ from $T^{\tilde{S}}$ and the conditional KL divergence of $T^{r(S)}$ from $T^{r(\tilde{S})}$ as 
\begin{equation}
\label{IToVI}
ITVoI=D_{KL}(T^{\tilde{S}}||T^{S}\otimes T^{I})+D_{KL}(T^{r(\tilde{S})}||T^{r(S)}),
\end{equation}
\noindent where
\begin{align}
\label{eq33}
&D_{KL}(T^{\tilde{S}}||T^{S}\otimes T^{I})=\int_{\tilde{S}_{k+1},\tilde{S}_{k},a_{k}}p\{\tilde{S}_{k+1},\tilde{S}_{k},a_{k}\} \IEEEnonumber \\
&\log\left(\frac{p\{\tilde{S}_{k+1}|\tilde{S}_{k},a_{k}\}}{p\{S_{k+1}|S_{k},a_{k}\}p\{I_{k+1}|S_{k},a_{k},I_{k}\}}\right), 
\end{align}
\noindent and
\begin{align}
\label{eq34}
&D_{KL}(T^{r(\tilde{S})}||T^{r(S)})=\int_{R_{k+1},\tilde{S}_{k},a_{k}}p\{R_{k+1},\tilde{S}_{k},a_{k}\} \IEEEnonumber \\
&\log\left(\frac{p\{R_{k+1}|\tilde{S}_{k},a_{k}\}}{p\{R_{k+1}|S_{k},a_{k}\}}\right). 
\end{align}
\end{mydef9}

Note that the KL divergence in \eqref{eq33} and \eqref{eq34} is a measure of the information lost when $T^{S}\otimes T^{I}$ and $T^{r(S)}$ are used for approximating $T^{\tilde{S}}$ and $T^{r(\tilde{S})}$, respectively. The ITVoI is $0$ if the transition of $S_{k}$ and the reward $R_{k+1}$ are independent of $I_{k}$. In this case, we know from Lemma 1 that the performance of the optimal policies of the augmented-state SSDP and the original SSDP are the same, and there is no need to include $I_{k}$ in $\tilde{S}_{k}$. On the other hand, a higher KL divergence value indicates that the transition of $S_{k}$ or the reward $R_{k+1}$ depend on $I_{k}$ to a larger extent, and thus the availability of $I_{k}$ is more important for accurately predicting the future state $S_{k+1}$ or predicting the reward $R_{k+1}$. In this case, including $I_{k}$ in $\tilde{S}_{k}$ will be more likely to improve the performance of the optimal policy. Therefore, the KL divergence is a suitable quantitative measure for the VoI.\par 

We would like to emphasize that although KL divergence is used to quantify ITVoI, the ultimate objective remains to evaluate the impact of information on control performance, as established in Lemma 1. This contrasts with related concepts in information theory, such as information entropy, which aim to quantify uncertainty itself without regard to its effect on outcomes.\par

\newtheorem{case7}[case]{Case}
\begin{case7}
Consider the vehicle-following control problem in Case 1. In order to quantify the benefit of making decisions when $acc_{i-1,k}$ is available to the control agent of vehicle $i$, the ITVoI can be calculated for $acc_{i-1,k}$ as
\begin{align}
\label{ex7}
ITVoI& =D_{KL}(T^{\tilde{S}_{i}}||T^{S_{i}}\otimes T^{W_{i}})+D_{KL}(T^{r(\tilde{S}_{i})}||T^{r(S_{i})}) \IEEEnonumber \\
&=\int_{S_{i,k},acc_{i-1,k},u_{i,k}}p\{S_{i,k},acc_{i-1,k},u_{i,k}\}  \IEEEnonumber \\ &\log\left(\frac{1}{p\{acc_{i-1,k}|S_{i,k},u_{i,k}\}}\right) .
\end{align}

From \eqref{ex7}, we can see that the ITVoI of $acc_{i-1,k}$ depends on the conditional probability $p\{acc_{i-1,k}|S_{i,k},u_{i,k}\}$, i.e., how well we can predict the acceleration $acc_{i-1,k}$ of predecessor $i-1$ given the driving status $S_{i,k}$ and control input $u_{i,k}$ of the following vehicle $i$? Consider the extreme case when $acc_{i-1,k}=g(S_{i,k},u_{i,k})$ is a deterministic function of $S_{i,k}$ and/or $u_{i,k}$, we thus have $p\{acc_{i-1,k}|S_{i,k},u_{i,k}\}=\mathbf{1}_{acc_{i-1,k}=g(S_{i,k},u_{i,k})}$. Here, the ITVoI of $acc_{i-1,k}$ is $0$, meaning there is no value in knowing $acc_{i-1,k}$. Note that this scenario is not possible in practice, since $acc_{i-1,k}$ is a function of $acc_{i-1,k-1}$ and $u_{i-1,k-1}$ according to \eqref{eq12}. The predecessor $i-1$ cannot determine its control input $u_{i-1,k-1}$ at control interval $k-1$ based on $S_{i,k}$ and/or $u_{i,k}$, whose values are not available until control interval $k$. In the other extreme case, if the predecessor $i-1$ makes driving decisions completely independently of the following vehicle $i$, we have $p\{acc_{i-1,k}|S_{i,k},u_{i,k}\}=p\{acc_{i-1,k}\}$. This means the ITVoI depends on the probability distribution of $acc_{i-1,k}$. For example, if $acc_{i-1,k}$ is a constant value, i.e., $p\{acc_{i-1,k}=c\}=1$, the ITVoI is $0$. \par
\end{case7}

\newtheorem{remark5}[remark]{Remark}
\begin{remark5}[Comparison between utility-based VoI and ITVoI]
Compared to the utility-based VoI, the ITVoI is generally easier to estimate when the model of the environment, i.e., the state transition probabilities and the reward function, is known. This advantage arises because ITVoI does not require deriving the optimal policies of the SSDPs with state $S_k$ and augmented state $\tilde{S}_k$. However, there are several significant limitations to ITVoI. Firstly, it requires the knowledge of the environment model, which might not be accurately available in many practical tasks. Secondly, among the four types of utility-based VoI, the ITVoI is essentially equivalent to the EVoMI, as both aim to predict the benefit of having additional information $I_k$ before making decisions in terms of long-term average performance. However, ITVoI does not cover the other three types of utility-based VoI, i.e., IVoMI, EVoII, and IVoII, thereby limiting its applicability.  Finally, the utility-based VoI captures the ultimate objective of information usage, which is to improve the control performance. Meanwhile, the ITVoI measures the merit of information for improving some intermediate outcomes that could contribute to the enhancement of the ultimate objective. Therefore, utility-based VoI is more intuitive compared to ITVoI.   
\end{remark5}

\section{VoI Estimation using DRL}  	
For the joint design of control and communication, VoI should be estimated by the control agents and subsequently utilized by the communication agents to guide the optimization of communication decisions. This section focuses on VoI estimation, while VoI-oriented optimization for communication agents is presented in Section VI.\par

In Section V, we have defined two types of VoI: utility-based VoI and ITVoI. When the environment model is known, the ITVoI can be directly derived from \eqref{IToVI}, \eqref{eq33}, and \eqref{eq34}. Otherwise, the distributions of next states and rewards must be estimated from data \cite{lei2022deep}. In the following, we focus on the estimation of utility-based VoI, including EVoI and IVoI.\par   

To obtain either EVoI or IVoI, we require both policies $\pi^{\rm{inf}}$ and $\pi^{\rm{sup}}_{k}$ in Definition 7 and Definition 8. These policies can be derived using RL, dynamic programming, or optimal control algorithms. The superior policy $\pi^{\rm{sup}}$ is typically learned in a simulated environment with full/perfect observations, while the inferior policy $\pi^{\rm{inf}}$ can be learned in an environment with either partial/imperfect or full/perfect observations. In the latter case, the partial/imperfect observations are treated as full/perfect during policy execution. \par  


\subsection{EVoI Estimation}  

After training the two policies, EVoI can be derived by testing the two policies for multiple episodes, deriving the performance of each policy in terms of the average cumulative rewards, and finally calculating the performance difference between the two policies. \par

\subsection{IVoI Estimation}  

The IVoI corresponds to the advantage function of the policy $\pi^{\rm{sup}}$ with full/perfect observations $S^{\rm{sup}}_k$, where various methods can be used to estimate the advantage function of a policy in RL. \par

\subsubsection{Method A} 
Based on the definition of IVoI in Definition 8, we can estimate it using the Monte Carlo method. Specifically, we begin by sampling initial states from their distribution and then generate multiple trajectories by executing three policies: $\pi^{\rm{inf}}$, $\pi^{\rm{sup}}$, and a mixed policy that randomly selects between $\pi^{\rm{inf}}$ and $\pi^{\rm{sup}}$ at each control interval.
Each trajectory comprises a sequence of states, where each state contains a pair of full/perfect and partial/imperfect observations, $S^{\rm{sup}}_k$ and $S^{\rm{inf}}_k$, respectively, at control interval $k$.\par 

From these trajectories, a subset of states is selected to form a \emph{rollout set}. For each state in this set, we sample two actions: $a^{\rm{inf}}_k$ from the acting policy $\pi^{\rm{inf}}$ based on $S^{\rm{inf}}_k$, and $a^{\rm{sup}}_k$ from the optimal policy $\pi^{\rm{sup}}$ based on $S^{\rm{sup}}_k$, respectively. Two roll outs are then performed starting from the state action pairs $(S^{\rm{sup}}_k,a^{\rm{inf}}_k)$ and $(S^{\rm{sup}}_k,a^{\rm{sup}}_k)$, respectively, with the optimal policy $\pi^{\rm{sup}}$ consistently used thereafter. \par

The difference in cumulative rewards between the two roll outs is calculated for each state in the rollout set. These differences can serve as target values for training an advantage function estimator, e.g., a deep neural network (DNN), using supervised learning.  \par

The main drawback of this method is that it requires a large amount of experience data to train the advantage estimator accurately. \par

\subsubsection{Method B}  
Since the IVoI is essentially the advantage function of the optimal policy $\pi^{\rm{sup}}$ under full/perfect observations,
it can be derived from the value function $v_{\pi^{\rm{sup}}}(S^{\rm{sup}}_{k})$ and Q function $q_{\pi^{\rm{sup}}}(S^{\rm{sup}}_{k},a_k)$ of $\pi^{\rm{sup}}$ according to \eqref{advantage}. Specifically, given $v_{\pi^{\rm{sup}}}(S^{\rm{sup}}_{k})$ and $q_{\pi^{\rm{sup}}}(S^{\rm{sup}}_{k},a_k)$, the IVoI can be computed as their difference 
\begin{align}
\label{ivoia}
\xi_k&=A_{\pi^{\rm{sup}} }(S_{k}^{\rm{sup}}, a_{k}^{\rm{inf}})|_{a_{k}^{\rm{inf}}\sim\pi^{\rm{inf}}(\cdotp|S_{k}^{\rm{inf}})} \IEEEnonumber \\&=q_{\pi^{\rm{sup}}}(S^{\rm{sup}}_{k},a_{k}^{\rm{inf}})|_{a_{k}^{\rm{inf}}\sim\pi^{\rm{inf}}(\cdotp|S_{k}^{\rm{inf}})}-v_{\pi^{\rm{sup}}}(S^{\rm{sup}}_{k}).
\end{align}

Using \eqref{ivoia}, we convert the advantage estimation problem into the task of estimating the value and Q functions of $\pi^{\rm{sup}}$. This can be achieved by training two separate estimators using a method similar to Method A. However, this approach is less appealing, as training two estimators is generally more cumbersome than training a single estimator for the advantage function directly. \par

Method B becomes more attractive in the case of deterministic policies, where the value function can be computed from the Q function as $v_{\pi^{\rm{sup}}}(S^{\rm{sup}}_{k})=q_{\pi^{\rm{sup}}}(S^{\rm{sup}}_{k},\pi^{\rm{sup}}(S^{\rm{sup}}_{k}))$, so only the Q function needs to be estimated. Classical DRL algorithms such as deep Q-network (DQN) \cite{mnih2015human} and deep deterministic policy gradient (DDPG) \cite{lillicrap2015continuous} already learn Q functions during training. This implies that  a Q function estimator is readily available when these algorithms are used to learn $\pi^{\rm{sup}}$. However, this estimator may be less accurate for actions preferred by the acting policy $\pi^{\rm{inf}}$ but rarely chosen by $\pi^{\rm{sup}}$. Therefore, it is advisable to fine-tune the Q function estimator using a Monte Carlo method similar to that used in Method A. \par

\subsubsection{Method C} 
Another method is to only train a single estimator for the value function $v_{\pi^{\rm{sup}}}(S^{\rm{sup}}_{k})$ and estimate the advantage function using the Temporal Difference (TD) error 
\begin{equation}
\delta_{\pi^{\rm{inf}},k}= R_{k+1} + \gamma v_{\pi^{\rm{sup}}}(S^{\rm{sup}}_{k+1}) - v_{\pi^{\rm{sup}}}(S^{\rm{sup}}_{k}).
\end{equation}

In this case, the IVoI can be estimated at each control interval based on the reward $R_{k+1}$, resulting from executing the action $a_{k}^{\rm{inf}}\sim (\cdotp|S^{\rm{inf}}_{k})$, and the value function $v_{\pi^{\rm{sup}}}$ of the optimal policy for both the current state $S^{\rm{sup}}_{k}$ and the next state $S^{\rm{sup}}_{k+1}$. Since 
\begin{align}
\label{eq31}
\xi_k=A_{\pi^{\rm{sup}} }(S_{k}^{\rm{sup}}, a_{k}^{\rm{inf}})|_{a_{k}^{\rm{inf}}\sim\pi^{\rm{inf}}(\cdotp|S_{k}^{\rm{inf}})} =\mathrm{E}_{S^{\rm{sup}}_{k+1}}[\delta_{\pi^{\rm{inf}},k}],
\end{align}
\noindent the TD error provides an unbiased estimate of the IVoI, assuming the value function estimator accurately approximate the true value function $v_{\pi^{\rm{sup}}}(S^{\rm{sup}}_{k})$. While this method introduces some noise, its variance is relatively low since the expectation in \eqref{eq31} is taken only over the next state $S^{\rm{sup}}_{k+1}$.  \par

This method is particularly attractive when DRL algorithms such as asynchronous advantage actor-critic (A3C) \cite{mnih2016asynchronous} and proximal policy optimization (PPO) \cite{schulman2017proximal} are used to learn $\pi^{\rm{sup}}$, as these algorithms inherently estimate the value function while training stochastic policies. However, similar to Method B, the value function estimator may be less accurate for states that are frequently visited by the acting policy $\pi^{\rm{inf}}$ but rarely encountered under the optimal policy $\pi^{\rm{sup}}$. Therefore, fine-tuning the value estimator using a Monte Carlo method — similar to Method A — is recommended.

\begin{figure}
	\centering
	\includegraphics[width=0.4\textwidth]{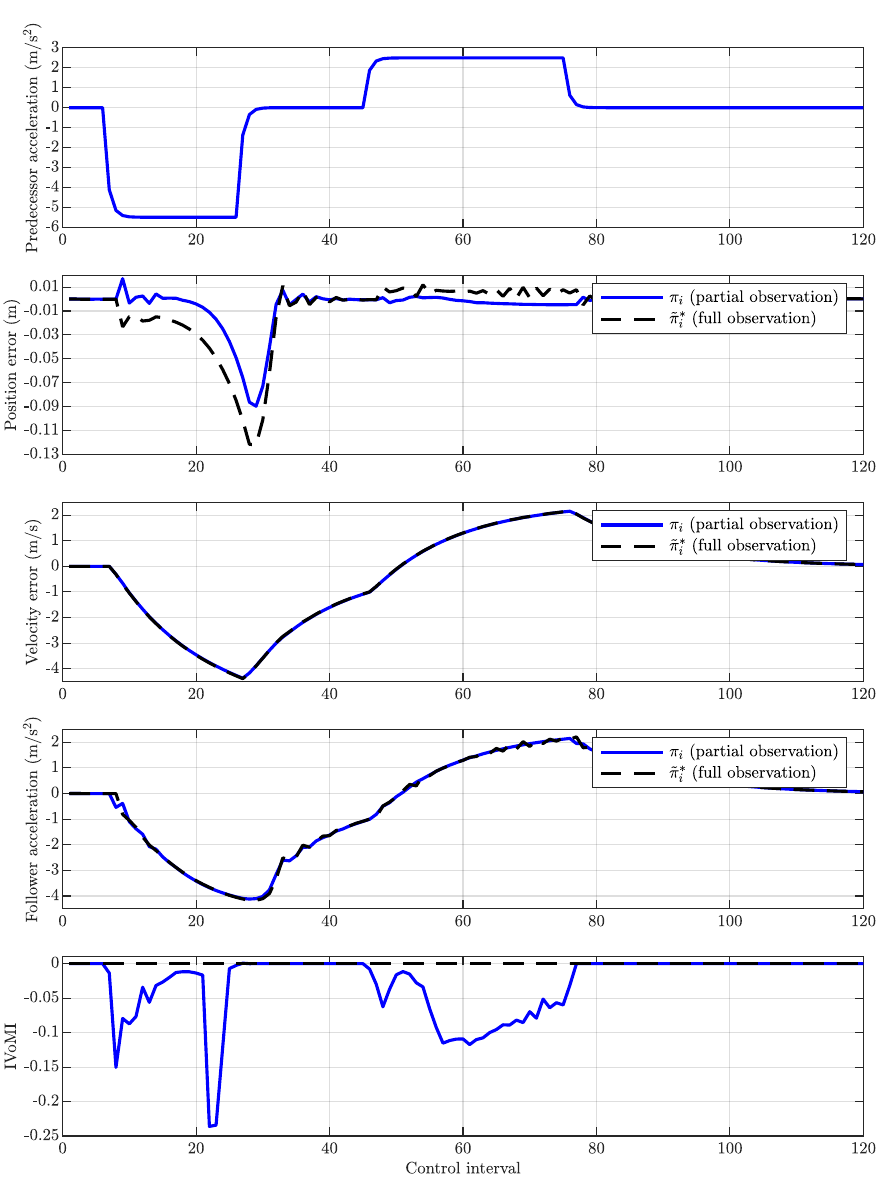}
	\caption{The IVoMI for the vehicle-following control problem in Case 6 during a specific test episode in the Stop-and-Go scenario, shown alongside the acceleration trajectory of the predecessor ($acc_{i-1,k}$), the position error ($e_{pi,k}$), velocity error ($e_{vi,k}$), and the follower's acceleration ($acc_{i,k}$) under two policies: $\tilde{\pi}_{i}^{*}$ and $\pi_{i}$.}
	\label{ivoi}
\end{figure}

\newtheorem{case8}[case]{Case}
\begin{case8}
We estimate EVoMI and IVoMI for the vehicle-following control problem as given in Case 6. The velocity trajectory of the preceding vehicle $i-1$ is extracted from the Next Generation Simulation (NGSIM) dataset \cite{NGSIM2009} when training the following vehicle. The optimal policy $\tilde{\pi}_{i}^{*}$ for the augmented-state SSDP in Case 4 is learned by the Twin Delayed Deep Deterministic policy gradient (TD3) algorithm \cite{fujimoto2018addressing}. This policy is also used as the acting policy $\pi_{i}$ when the predecessor's acceleration is not available. In this case, $acc_{i-1,k}$ is set to $0$ in the state. \par 

The EVoMI is $-3.12$, which is calculated by evaluating the policy over $100$ test episodes in which $acc_{i-1,k}$ is either consistently unavailable or consistently available throughout the entire episode, and then computing the difference in the average total reward between the two settings. \par 

Fig. \ref{ivoi} shows the IVoMI during a specific test episode in the Stop-and-Go scenario, along with the acceleration trajectory of the predecessor ($acc_{i-1,k}$), the position error ($e_{pi,k}$), velocity error ($e_{vi,k}$), and follower's acceleration ($acc_{i,k}$) under the two policies: $\tilde{\pi}_{i}^{*}$ and $\pi_{i}$. \par 

At each control interval $k$, two actions are selected: one by the optimal policy $\tilde{\pi}_{i}^{*}$ when $acc_{i-1,k}$ is available, and the other by the acting policy $\pi_{i}$ when $acc_{i-1,k}$ is not available, respectively. The action prescribed by the acting policy is executed to determine the next state. Meanwhile, the advantage function — evaluated by the trained estimator using the full state, including $acc_{i-1,k}$, and the action selected by the acting policy $\pi_{i}$ as inputs — is used to quantify the IVoMI. \par

Note that the EVoMI is a constant negative value as it does not depend on specific state action pairs. This negative value indicates that transmitting the predecessor's acceleration to the follower is beneficial for vehicle control. In contrast, the IVoI varies across control intervals since it depends on the state action pairs. When $acc_{i-1,k}=0$, IVoMI is zero, indicating the information is not useful in this case. Otherwise, the IVoMI takes on a negative value. 
\end{case8}

\section{VoI-Oriented Optimization in Communication Systems}
When full and perfect information corresponding to the environment state is available, the control agent can typically make decisions that yield better performance than those based on partial and/or imperfect observations. However, radio resources in wireless networks are limited and valuable. Therefore, the communication agents have to make decisions on information transmission and RRM in order to strike a good tradeoff between minimizing control performance degradation due to non-ideal communication and maximizing communication performance for traditional services. \par

\subsection{Three Types of Communication Decisions}

The communication decisions can be divided into three broad categories as explained below.\par

  \subsubsection{What to Communicate}
	The remote sensory information normally consists of multiple elements. The communication agents can make decisions on
	
	\begin{itemize}
		\item the subset of information that should be transmitted;
		\item the processed information for transmission.
	\end{itemize}

	\subsubsection{When to Communicate}
	It might not be necessary to transmit a piece of (processed) information at every control interval, depending on the specific instance of information. Therefore, the communication agent can make decisions on whether the element of information should be transmitted or not at each control interval based on its content and relevance with the previous information. \par

	\subsubsection{How to Communicate}
	For a piece of (processed) information to be transmitted at a control interval, the radio resources need to be scheduled to support its transmission. Considering the stringent resource limitations, the communication agents can make decisions to schedule available resources for transmission of the most meritorious information. \par

\subsection{Three Time Scales for Communication Decision-Making}

\subsubsection{Every Communication Interval - How to Communicate}
Since RRM decisions are normally made at a faster time scale (e.g., every $1$ ms in fourth-generation (4G) cellular system) than the control decisions, one control interval can be further divided into $T$ communication intervals. Let $\Delta t$ represent the duration of a communication interval. We denote the $t$-th communication interval in control interval $k$ as communication interval $(k,t)$, where $t\in\mathcal{T}=\{0,1,2,\cdots,T-1\}$. The ``How to Communicate" decisions are made at every communication interval. \par

\subsubsection{Every Control Interval - When and What to Communicate}
Since sensory information is sampled at the beginning of each control interval, the ``When to Communicate" and ``What to Communicate" decisions can be made dynamically per control interval. The communication agent determines whether to communicate with its peers and, if so, selects the subset of information or processed information to share during this control interval. \par

\subsubsection{Across Multiple Control Intervals - When and What to Communicate}
The ``When to Communicate" and ``What to Communicate" decisions can also be made statically only once at control interval $0$ or semi-statically for every $Z>1$ control intervals. In this case, the subset of information or processed information will always be transmitted at every control interval within the effective time range of the decision. An example is choosing the information topology for a platoon \cite{lei2022deep}. \par 

  	\newtheorem{case9}[case]{Case}
   	\begin{case9}
  		We consider a typical urban C-V2X network, where V2V links coexist with V2I links. A V2I link connects a vehicle to the BS and can be used for high-throughput services. A V2V link connects a pair of preceding vehicle $i-1$ and following vehicle $i$ for periodic transmission of cooperative awareness messages (CAM) to support advanced driving services. Specifically, at the beginning of each control interval $k$, the acceleration $acc_{i-1,k}$ of the predecessor $i-1$ is sampled to form the CAM and buffered in a queue before transmission to the corresponding follower $i$. \par
    
    At each communication interval $(k,t)$, the predecessor $i-1$ transmits the CAM generated at the beginning of each control interval $k$. Any CAM that is not fully transmitted by the end of the last communication interval $(k,T-1)$ of control interval $k$ will be discarded. In other words, any CAM not fully delivered during the previous control interval $k-1$ is replaced by newly generated CAM at the beginning of control interval $k$.\par

    As introduced in Case 1, the control agent in follower $i$ determines its control input based on the local sensory information $ep_{i,k}$, $ev_{i,k}$, $acc_{i,k}$, as well as the remote sensory information received from its predecessor $i-1$. \par
        
        There is a communication agent in each predecessor $i-1$ to make communication decisions, which fall into one or more of the following categories:
         \begin{itemize}
    	\item Every Communication Interval - How to Communicate: The communication agent makes RRM decisions at each communication interval $(k,t)$. To enhance spectrum utilization, we consider a setting where $L$ V2V links share the uplink resources orthogonally allocated to $M$ V2I links. With a slight abuse of notation, each V2V link, connecting a predecessor $i-1$ and its corresponding follower $i$, is indexed by $i-1 \in \mathcal{L} = \{1, 3, 5, \ldots, 2L-1\}$. Without loss of generality, we assume that every V2I link $m\in \mathcal{M}=\{0,\cdots, M-1\}$ is pre-assigned sub-channel $m$ and transmits with a constant power $P^{\rm I}_{m}$. Moreover, one or more V2V links can reuse the sub-channels of the V2I links for CAM transmission.  We use the binary allocation indicator $\theta_{i-1,m,(k,t)}\in\{0,1\}$ to indicate whether V2V link $i-1$ occupies sub-channel $m$ at communication interval $(k,t)$ or not. Moreover, we consider that each V2V link $i-1$ occupies at most one sub-channel, i.e., $\sum_{m=0}^{M-1} \theta_{i-1,m,(k,t)} \leq 1$. Let $P^{\rm V}_{i-1,m,(k,t)}$ denote the transmit power of V2V link $i-1$ over the sub-channel $m$ at communication interval $(k,t)$. Therefore, each communication agent $i-1$ has to make the subchannel allocation $\{\theta_{i-1,m,(k,t)}\}_{m\in\mathcal{M}}$ and power allocation $\{P^{\rm V}_{i-1,m,(k,t)}\}_{m\in\mathcal{M}}$ decisions at each communication interval $(k,t)$.
        \item Every Control Interval - When to Communicate: Let $\phi_{i-1,k}\in\{0,1\}$ be the binary transmission indicator to indicate whether the sampled CAM $acc_{i-1,k}$ will be transmitted to the follower $i$ or not. At each control interval $k$, each communication agent $i-1$ can make decisions on $\phi_{i-1,k}$. 
        \item Across Multiple Control Intervals - When to Communicate: Each communication agent $i-1$ makes a one-time decision on whether it will transmit its acceleration or not to its follower $i$. Once the decision is made, the predecessor will always transmit $acc_{i-1,k}$ or not at each control interval $k$.
        \end{itemize}
    	\end{case9}

\subsection{Optimization Objective}
In general, the optimization objective is to maximize communication performance for traditional services, while minimizing the control performance degradation in vehicle control due to non-ideal communications, i.e.,
\begin{align}
\label{RRAObjective}
\max_{\pi_{\rm CM}} J^{\mathrm{CM}} =   \max_{\pi_{\rm CM}} 
( \kappa_1 J^{\mathrm{CM\_trad}} + \kappa_2 \Bar{\Xi}),
\end{align}
where \(J^{\mathrm{CM\_trad}}\) represents the long-term average communication performance for traditional services, such as the average throughput, and \(\Bar{\Xi}\) is the EVoI. The weighting coefficients \(\kappa_1\) and \(\kappa_2\) reflect the relative importance of traditional network services and control services in the optimization objective.\par

According to Lemma 2, the optimization objective in \eqref{RRAObjective} can be expressed in the following form of expectation.
\begin{align}
&J^{\mathrm{CM}}  \IEEEnonumber \\&=\mathrm{E}_{\pi^{\rm CM}}\sum_{k=0}^{\infty}\left[\kappa_1\left(\sum_{t=0}^{T-1} \gamma^{kT+t}r_{(k,t)}^{\mathrm{CM\_trad}}\right)+\kappa_2\gamma^{k}\mathrm{E}_{\pi^{\rm inf}}[\xi_{k+1}] \right],
\end{align}
\noindent where $r_{(k,t)}^{\mathrm{CM\_trad}}$ represents the instantaneous communication performance at communication interval $(k,t)$, e.g., instantaneous data rate, where 
\begin{equation}
J^{\mathrm{CM\_trad}}=\mathrm{E}_{\pi^{\rm CM}}\left[\sum_{k=0}^{\infty}\sum_{t=0}^{T-1} \gamma^{kT+t}r_{(k,t)}^{\mathrm{CM\_trad}}\right],
\end{equation}
\noindent and $\xi_{k+1}$ is the IVoI. \par

\newtheorem{case10}[case]{Case}
   	\begin{case10}
    In Case 9, the optimization objective of the communication agents $i-1\in\mathcal{L}$ is to maximize the sum V2I throughtput while minimizing the performance degradation of the followers $i$ due to non-ideal communications. \par
    
    The instantaneous channel gain of V2V link $i-1$ over sub-channel $m$ (occupied by V2I link $m$) at communication interval $(k,t)$ is denoted by $G_{i-1,m,(k,t)}$. $G_{i-1,m,(k,t)}$ remains constant within a communication interval, and can be written as
    \begin{equation}
    G_{i-1,m,(k,t)}=\alpha_{i-1,(k,t)}h_{i-1,m,(k,t)},
    \end{equation} 
   \noindent where $\alpha_{i-1,(k,t)}$ captures the large-scale fading effects including path loss and shadowing, which are assumed to be frequency independent. $h_{i-1,m,(k,t)}$ corresponds to frequency-dependent small-scale fading.\par 
    
    Similarly, let $G_{m,(k,t)}$ denote the instantaneous channel gain of V2I link $m$; $G_{i-1,B,m,(k,t)}$ the interference channel gain from V2V link $i-1$ transmitter at predecessor $i-1$ to V2I link $m$ receiver at the BS; $G_{B,i-1,m,(k,t)}$ the interference channel gain from V2I link $m$ transmitter at a vehicle to V2V link $i-1$ receiver at follower $i$; and $G_{j-1,i-1,m,(k,t)}$ the interference channel gain from V2V link $j-1$ transmitter at predecessor $j-1$ to V2V link $i-1$ receiver at follower $i$ over the sub-channel $m$.\par

    The SINR $\gamma_{m,(k,t)}$ of V2I link $m$ and the SINR $\gamma_{i-1,m,(k,t)}$ of V2V link $i-1$ on sub-channel $m$ at communication interval $(k,t)$ are derived by 
    \begin{align}\label{SINRV2I}
    & \gamma_{m,(k,t)} = \IEEEnonumber \\ 
    &\frac{P^{\rm I}_{m} G_{m,(k,t)}}{\sigma^2 + \sum\limits_{i-1\in\mathcal{L}}\phi_{i-1,k}\theta_{i-1,m,(k,t)} P^{\rm V}_{i-1,m,(k,t)} G_{i-1,B,m,(k,t)}},
    \end{align}
    and
    \begin{align}\label{SINRV2V}
    \gamma_{i-1,m,(k,t)}= \frac{P^{\rm V}_{i-1,m,(k,t)} G_{i-1,m,(k,t)}}{\sigma^2 + I_{i-1,m,(k,t)}},
    \end{align}
    \noindent respectively. $\sigma^2$ is the power of channel noise which satisfies the independent Gaussian distribution with a zero mean value. $I_{i-1,m,(k,t)}$ in \eqref{SINRV2V} is the total interference power received by V2V link $i-1$ over sub-channel $m$, where
    \begin{align}\label{IV2V}
    & I_{i-1,m,(k,t)}=\IEEEnonumber \\
    & P^{\rm I}_{m} G_{B,i-1,m,(k,t)}+ \IEEEnonumber \\
    & \sum\limits_{j-1\in \mathcal{L} \backslash \{i-1\}}\phi_{i-1,k}\theta_{j-1,m,(k,t)}  P^{\rm V}_{j-1,m,(k,t)} G_{j-1,i-1,m,(k,t)}.\IEEEnonumber
    \end{align}
    
     The instantaneous data rates $C_{m,(k,t)}$ and $C_{i-1,(k,t)}$ of V2I link $m$ and V2V link $i-1$ at communication interval $(k,t)$ are respectively derived as
    \begin{equation}
    \label{rateV2I}
    C_{m,(k,t)} = B\log_2(1+\gamma_{m,(k,t)}),
    \end{equation}
    and
    \begin{align}\label{rateV2V}
    C_{i-1,(k,t)} =\sum_{m=0}^{M-1} {\theta_{i-1,m,(k,t)}} B\log_2(1+\gamma_{i-1,m,(k,t)}),
    \end{align}
    \noindent where $B$ is the bandwidth of a sub-channel.\par
    Let $C^{\rm CAM}_{i-1,(k,t)}$ denote the transmission rate of V2V link $i-1$ in terms of CAM at communication interval $(k,t)$, which is given by
    \begin{equation}
    \label{rateV2VCAM}
    C^{\rm CAM}_{i-1,(k,t)}=\frac{C_{i-1,(k,t)}}{N_c},
    \end{equation}
    \noindent where $N_c$ is the constant CAM size. \par

       Each predecessor $i-1\in\mathcal{L}$ maintains a buffer to store its CAMs. Let $q^{\rm CAM}_{i-1,(k,t)}$ denote the queue length, i.e., the number of CAMs in the buffer of predecessor $i-1$ at communication interval $(k,t)$. At the beginning of the first communication interval $(k, 0)$ of each control interval $k\in\mathcal{K}$, predecessor $i-1$ samples the driving status for control interval $k$ and generates a CAM. If the communication agent decides to transmit this CAM ($\phi_{i-1,k}=1$), it is stored in the buffer. During each communication interval $(k, t)$, where $t\in\mathcal{T}$, the queue length $q^{\rm CAM}_{i-1,(k,t)}$ is decreased by $\Delta t\times C^{\rm CAM}_{i-1,(k,t)}$, which represents the number of CAMs transmitted during this communication interval.\par 
       
       Since any CAM not fully transmitted by the end of the last communication interval $(k,T-1)$ is discarded, the queue is reset at the start of each new control interval $k+1$. Specifically, at communication interval $(k+1, 0)$, $q^{\rm CAM}_{i-1,(k+1,0)}$ is set to $1$ if a CAM is scheduled for transmission ($\phi_{i-1,k+1}=1$), or $0$ otherwise ($\phi_{i-1,k+1}=0$). Because the queue length changes abruptly at the boundaries between control intervals, we use $q^{\rm CAM}_{i-1,(k,T)}$ to denote the queue length at the end of control interval $k$, i.e., after the final communication interval $(k,T-1)$. Note that $q^{\rm CAM}_{i-1,(k,T)}\neq q^{\rm CAM}_{i,(k+1,0)}$ due to this discontinuity. \par
  
    The queue process evolves as
    	\begin{align}
    	\label{queue2}
    	\setlength{\arraycolsep}{1.6pt}
    	& q^{\rm CAM}_{i-1,(k,t)}= \IEEEnonumber \\
    	&\left\{
    	\begin{array}{ll}
    	\phi_{i-1,k} , &\mathrm{if}\quad t=0 \\
    	\max \left[0,q^{\rm CAM}_{i-1,(k,t-1)}- C^{\rm CAM}_{i-1,(k,t-1)}\Delta t\right], & \mathrm{if} \quad 0<t\leq T \\
    	\end{array}\right. .
    	\end{align}
    	
    For each follower $i$, if a CAM is fully delivered by the end of control interval $k$, the receiver will update its stored CAM with the newly received one at the beginning of the next control interval $k+1$ and uses the updated information for vehicle control. If a CAM is not received or only partially received, the receiver has two option: (1) replace the missing CAM with dummy bits (e.g., zeros); (2) continue using the last fully received CAM. \par 
    
    With option (2), when the most recent available CAM of follower $i$ at the beginning of control interval $k$ corresponds to the CAM sampled at the predecessor $i-1$ in control interval $k-\tau_{i,k}$, we refer to $\tau_{i,k}$ as the observation delay at V2V receiver $i$, which is derived from the queue length as
    	\begin{align}
    	\label{aoi2}
    	\tau_{i,k}=	\left\{
    	\begin{array}{ll}
    	1 , &\mathrm{if}\quad 	q^{\rm CAM}_{i-1,(k,T)}=0 \\
    	\tau_{i,k}+1, & \mathrm{otherwise} \\
    	\end{array}\right. .
    	\end{align}

      The optimization objective is formulated as 
\begin{equation}
\label{opti_obj}
J^{\mathrm{CM}}=\kappa_1 \sum_{m=0}^{M-1}\mathrm{E}_{\pi^{\rm CM}}\left[\sum_{k=0}^{K-1}\sum_{t=0}^{T-1} \gamma^{kT+t}C_{m,(k,t)}\right]+\kappa_2 \sum_{i-1\in\mathcal{L}}\Xi_{i},
\end{equation}
\noindent where $\gamma$ is the communication reward discount factor. The first term corresponds to the discounted sum throughput of all V2I links $m\in\mathcal{M}$ over all the communication intervals of a control episode, and the second term corresponds to the EVoMI $\Xi_{i}^{\rm M}$ or EVoII $\Xi_{i}^{\rm I}$ of follower $i$ defined in Case 6. The weight factors $\kappa_1$ and $\kappa_2$ indicate the relative importance of minimizing control performance degradation versus maximizing the V2I throughput.\par

The optimization objective $J^{\mathrm{CM}}$ should correspond to the expected return of the DRL model, i.e.,
\begin{equation}
\label{RRA_cumreward}
J^{\mathrm{CM}}=\mathrm{E}_{\pi^{\rm CM}}\mathrm{E}_{\pi^{\rm inf}}\left[\sum_{k=0}^{K-1}\sum_{t=0}^{T-1} \gamma^{kT+t} r^{\rm CM}_{(k,t)}\right], 0 \leq \eta \leq 1,
\end{equation}
\noindent where the expectation is taken with respect to the probability distribution of the state-action trajectories when the communication policy is $\pi^{\rm CM}$ and vehicle control policy is $\pi^{\mathrm{inf}}$. \par

	\end{case10}

\subsection{VoI-oriented Optimization for Vehicular Communication}

When addressing the problem of ``When and What to Communicate", different approaches can be adopted depending on the decision-making time scale: 
  \begin{itemize}

   \item \textit{Across Multiple Control Intervals}: When decisions are made statically or semi-statically across multiple control intervals, one can evaluate $J^{\mathrm{CM}}$ under different communication decisions and choose the decision that maximizes $J^{\mathrm{CM}}$. To this end, the EVoI can be estimated as outlined in Section V, and $J^{\mathrm{CM\_trad}}$ can be evaluated using analytical tools such as queuing theory or through simulation.

    	\item  \textit{Every Control Interval}: 
        When decisions are made dynamically at each control interval, an SSDP can be formulated to optimize  communication decisions at each step. In this case, the reward function of the SSDP can be expressed as
\begin{align}
\label{com_reward_1}
r^{\rm CM}_{k}= \kappa_1\left(\sum_{t=0}^{T-1} \gamma^{t}r_{(k,t)}^{\mathrm{CM\_trad}}\right)+\kappa_2\xi_{k+1},
\end{align}
\noindent where \(\xi_{k+1} = A_{\pi^{\text{sup}}}(S_{k+1}^{\text{sup}},a_{k+1})|_{a_{k+1}^{\text{inf}}\sim\pi^{\text{inf}}(\cdot|S_{k+1}^{\text{inf}})}\) is the IVoI. This terms captures the expected control performance degradation the recipient may experience due to making a decision at the next control interval \(k+1\) based on the partial/imperfect observation \(S_{k+1}^{\text{inf}}\) rather than the full/perfect observation \(S_{k+1}^{\text{sup}}\). \par

The partial/imperfect observation available at control interval \(k+1\) is affected by the ``When and What to Communicate" decisions made by the communication agent at control interval $k$. Specifically,
\begin{itemize}
    \item If $\xi_{k+1}$ corresponds to the IVoMI, it helps determine whether the raw information generated at control interval $k$ should be transmitted. 
    \item If $\xi_{k+1}$ corresponds to the IVoII, it guides the selection of which processed information should be shared.
\end{itemize}
        
  \end{itemize}

For the ``How to Communicate" problem, an SSDP can be formulated to optimize the communication decisions made at each communication interval. Specifically, the reward function of the SSDP can be expressed as
\begin{align}
\label{com_reward}
r^{\rm CM}_{(k,t)}=\left\{
\begin{array}{ll}
\kappa_1 r_{(k,t)}^{\mathrm{CM\_trad}}, &  \ 0 \le t< T-1\\
\kappa_1 r_{(k,t)}^{\mathrm{CM\_trad}}+\kappa_2 \xi_{k+1}, & \  t= T-1 \\
\end{array}\right. .
\end{align}
\noindent Note that the \(r_{(k,t)}^{\mathrm{CM\_trad}}\) component is present in the reward of every communication interval, while the IVoII component, i.e., \(\xi_{k+1}\), is only present in the last communication interval of each control interval when \(t = T-1\). The imperfect observation received by the control agent at control interval \(k+1\) is affected by the degree of communication nonideality the transmitted data has experienced, which is in part a result of the sequence of decisions on communication actions \(a_{(k,0)}^{\text{CM}}, a_{(k,1)}^{\text{CM}}, \ldots, a_{(k,T-1)}^{\text{CM}}\) that are selected during control interval \(k\) by the communication agent. \par

Since EVoI equals the expected cumulative IVoI according to Lemma 2, the expected discounted sum of the immediate rewards in \eqref{com_reward_1} and \eqref{com_reward} equal the optimization objective in \eqref{RRAObjective}.  

\begin{figure}
	\centering
	\includegraphics[width=0.4\textwidth]{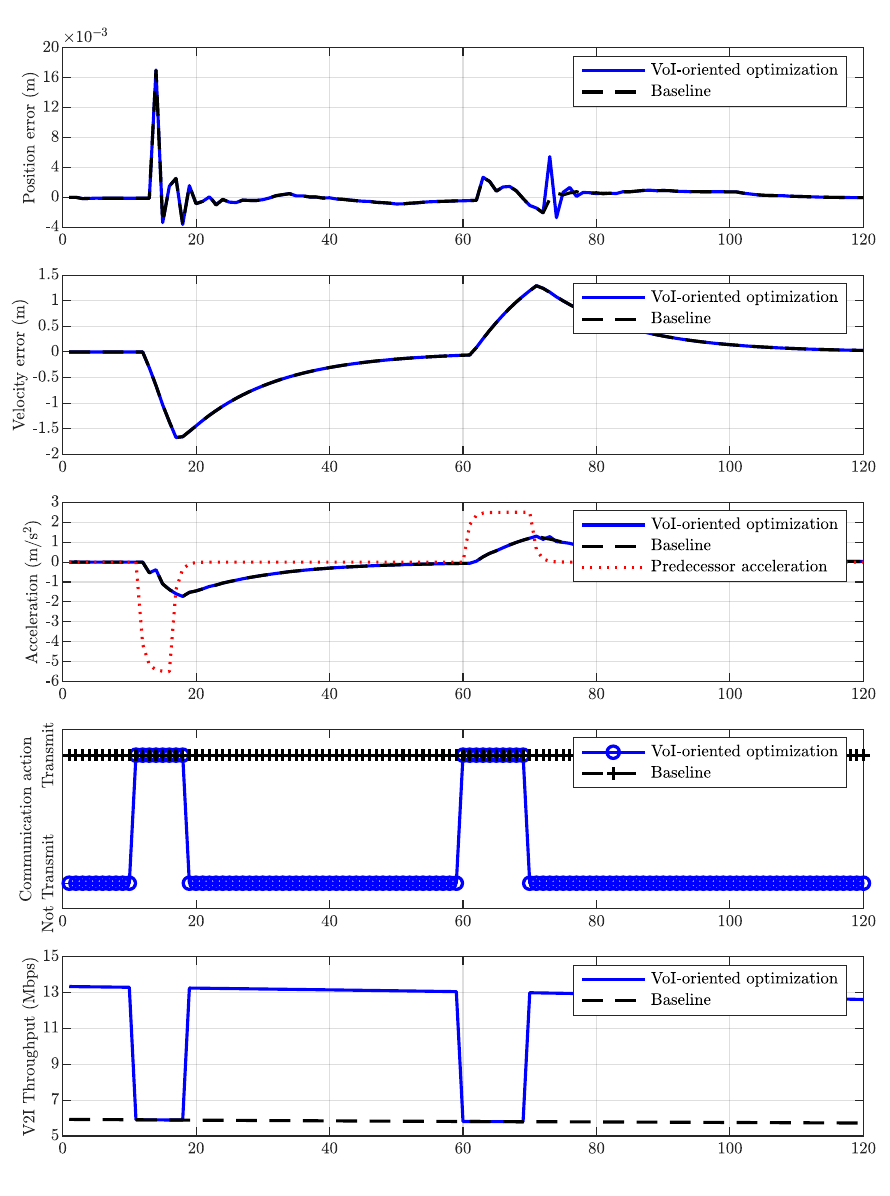}
	\caption{Performance comparison between the VoI-oriented optimization approach and a baseline method that always transmits the predecessor's acceleration, evaluated in a simplified C-V2X network supporting vehicle-following control applications.}
	\label{commun}
\end{figure}

 \newtheorem{case11}[case]{Case}
   \begin{case11}
The state space, action space, and reward function of the SSDPs for making communication decisions in Case 10 are given below. The SSDPs can be solved by a centralized communication agent at the BS; or by the decentralized communication agents at each predecessor $i-1\in\mathcal{L}$. The algorithms to solve the SSDP models are out of the scope of this paper. We refer the readers to \cite{10376461} for more details on a similar example.  
\begin{itemize}
 \item State Space
  \begin{itemize}
  \item When to Communicate: 
  
  $S^{\rm CM}_{(k)}=\{\boldsymbol G_{(k,0)},u_{i-1,k}\}$
  \item How to Communicate: 
  
  $S^{\rm CM}_{(k,t)}=\{\boldsymbol G_{(k,t)},q^{\rm CAM}_{(k,t)},\{u_{i-1,k'}\}_{k'=k-\tau_{\mathrm{max}}+1}^{k},t\}$,
  \end{itemize}
 \item Action Space
 \begin{itemize}
 \item When to Communicate: 
 
 $a^{\rm CM}=\{\phi_{i-1,k}\}_{i-1\in\mathcal{L}}$
 \item How to Communicate: 
 
 $a^{\rm CM}=\{\theta_{i-1,m,(k,t)},P^{\rm V}_{i-1,m,(k,t)}\}_{m\in\mathcal{M},i-1\in\mathcal{L}}$.
 \end{itemize}
 \item Reward Function
 \begin{itemize}
  \item When to Communicate: 
\begin{align}
\label{com_reward_1_ex}
r^{\rm CM}_{(k,t)}&= \kappa_1\left(\sum_{m=0}^{M-1}\sum_{t=0}^{T-1} \gamma^{t}C_{m,(k,t)}\right) \IEEEnonumber \\
    & +\kappa_2\sum_{i-1\in\mathcal{L}}\xi_{i,k+1}.
\end{align}
   \item How to Communicate:
 \begin{align}
\label{com_reward_ex}
& r^{\rm CM}_{(k,t)}= \IEEEnonumber \\ 
& \left\{
\begin{array}{ll}
\kappa_1 \sum_{m=0}^{M-1}C_{m,(k,t)}, &  \ 0 \le t< T-1\\
\kappa_1 \sum_{m=0}^{M-1}C_{m,(k,t)}+ & \\ \kappa_2 \sum_{i-1\in\mathcal{L}}\xi_{i,k+1}, & \  t= T-1 \\
\end{array}\right. .
\end{align}
  \end{itemize}
\end{itemize}

To demonstrate the advantages of VoI-oriented optimization, we conduct experiments on a simplified C-V2X network supporting vehicle-following control applications, as considered in Case 9 and Case 10, with $M=1$ V2I link and $L=1$ V2V link. Focusing on the ``When to Communicate" problem, we consider that if a CAM is scheduled for transmission, the predecessor always transmits at maximum power.  \par

Fig. \ref{commun} shows a performance comparison between the VoI-oriented optimization approach and a baseline method that always transmits the predecessor's acceleration. The VoI-oriented approach selectively suppresses CAM transmissions when the predecessor's acceleration is zero, recognizing that such information holds no value for the follower, as indicated by the IVoI shown in Fig. \ref{commun}. As a result, the V2I link experiences less interference from the V2V link during these periods, resulting in higher throughput compared to the baseline. Meanwhile, the position error ($e_{pi,k}$), velocity error ($e_{vi,k}$), and follower's acceleration trajectories ($acc_{i,k}$) remain nearly identical under both methods, demonstrating that omitting the transmission of zero acceleration has a negligible impact on control performance.
    \end{case11} \par

\section{Conclusion and Future Work}
The significance of this paper is to establish a canonical framework for modeling the value of the information shared over 6G V2X networks, specifically for AD decisions under various forms of uncertainty. To lay the theoretical groundwork, we have presented the SSDP model as an extension of the MDP model, explicitly accounting for useful
information that may be missing from the state. This extension accommodates both fully observable and partially observable scenarios. Based on the SSDP model, we have proposed a mathematically rigorous and practical method to identify the set of information that can enhance decision-making if made available.\par

At the heart of this framework are two interrelated definitions of VoI — EVoI and IVoI — derived from the SSDP model. These definitions correspond to the regret of a policy with partial or imperfect observations and the advantage function of a policy with full or perfect observations, respectively, in the context of RL terminology. \par

The primary purpose of VoI modeling is to enable the integrated design and joint optimization of control and communication. Since VoI should be estimated by the control agent of the information recipient and shared with the communication agent of the information source to guide optimization of communication decisions, we have provided methods for estimating VoI using DRL, and for formulating the SSDP model to optimize communication decisions based on the estimated VoI. We have used a simple vehicle-following control problem as an illustrative example throughout the paper. \par

Although the framework has been developed with CAV systems in mind, it holds great promise for enabling the joint optimization of stochastic, sequential control and communication decisions across diverse networked control systems (NCS) and autonomous/artificial Internet of Things (AIoT) applications \cite{lei2020deep}, including networked robots, smart grid, industrial automation systems, unmanned aerial vehicle (UAV) swarms, and more. Several key challenges present compelling directions for future research. \par

Firstly, accurate VoI estimation by control agents is crucial for providing precise reward signals that guide the optimization of communication agents' decision-making. The estimation methods discussed in this paper, particularly those for IVoI, still leave room for improvement. This is because IVoI is essentially an advantage function, and existing estimation methods for advantage functions are primarily designed to support policy learning, rather than to achieve high-precision value estimates. As a result, the precision of the learned values - especially for those states and actions that are infrequently visited under policies with full or perfect observations - may be insufficient for accurate VoI estimation. One potential remedy is to combine Method A with Method B or Method C, aiming to achieve high estimation accuracy with a reasonable amount of experience.\par 


Secondly, the construction of communication state variables in SSDP models plays a critical role in the performance of the learned communication policies. Generally speaking, the communication state should comprise two key components. The first includes Channel State Information (CSI), Queue State Information (QSI), and other elements typically defined in MDP models for communication problems. The second component pertains to IVoI, which theoretically encompasses any information that is useful in predicting the IVoI at the recipient. The IVoI, as an advantage function, depends on the recipient's full or perfect observations, the action determined by the policy under partial or imperfect observations, and the neural network parameters for the critic of the full/perfect- observations policy.\par 

Ideally, the communication states should incorporate all of the above information. However, in practice, the composition of communication states varies across tasks, depending on what information is available to the communication agents. Developing a systematic method for constructing communication state representations that is agnostic to specific control tasks remains an open issue.  \par

Lastly, VoI appears only in the reward of the last communication interval within each control interval in \eqref{com_reward_ex}, resulting in a sparse reward problem. This issue is further exacerbated in multi-agent RL settings, which are common in RRM problems, due to increased complexity and coordination challenges among agents. Therefore, learning efficient communication policies under sparse reward conditions remains a critical issue to address. \par

\appendix
\subsection{Proof of Theorem 1}
When the SSDP in Definition 1 is an MDP, we have 
\begin{equation}
\label{aeq11}
\mathrm{Pr}(S_{k+1}| S_k, a_k) =\mathrm{Pr}(S_{k+1}| S_k, a_k, \{S_{k'},a_{k'}\}_{k'\leq k-1}).
\end{equation}

Therefore, from \eqref{aeq11}, we have
\begin{align}
\label{aeq12}
&\mathrm{Pr}(f^{S}(S_{k},a_{k},f^{W}(S_{k},a_{k},\zeta_{k}))| S_k, a_k) \IEEEnonumber  \\
=&\mathrm{Pr}(f^{S}(S_{k},a_{k},f^{W}(S_{k},a_{k},\zeta_{k}))| S_k, a_k, \{S_{k'},a_{k'}\}_{k'\leq k-1}),
\end{align}
\noindent which follows from the system's state transition function $S_{k+1}=f^{S}(S_{k},a_{k},W_{k})$ given in Definition 1. In addition, $W_{k}=f^{W}(\tilde{W}_{k})$ according to the exogenous information function in Definition 1, while $\tilde{W}_{k}\subseteq\{S_{k},a_{k},\zeta_{k}\}$ as stated in Theorem 1.\par

From \eqref{aeq12}, we have
\begin{align}
\label{aeq13}
&\mathrm{Pr}(\zeta_{k}| S_k, a_k) \IEEEnonumber  \\
=&\mathrm{Pr}(\zeta_{k} | S_k, a_k, \{S_{k'},a_{k'}\}_{k'\leq k-1}) \IEEEnonumber  \\
=&\mathrm{Pr}(\zeta_{k} | S_k, a_k, \{S_{k'},a_{k'},W_{k'}\}_{k'\leq k-1}),\IEEEnonumber  \\
\end{align}
\noindent where the last equation follows from $W_{k'}=f^{\rm WS}(S_{k'+1},S_{k'},a_{k'})$ since $S_{k'+1}=f^{\rm S}(S_{k'},a_{k'},W_{k'})$, and the property of conditional probability $\mathrm{Pr}(X|g(Y),Y)=\mathrm{Pr}(X|Y)$.

Since $W_{k}=f^{W}(S_{k},a_{k},\zeta_{k})$, $\zeta_{k}$ is a set of random variables whose distribution is independent of $S_k$ and $a_k$ by the expression of the formula. Thus, from \eqref{aeq13}, we have
\begin{align}
\label{aeq13a}
\mathrm{Pr}(\zeta_{k}) 
=\mathrm{Pr}(\zeta_{k} | \{S_{k'},a_{k'},W_{k'}\}_{k'\leq k-1}),
\end{align}
\noindent which shows that the distribution of $\zeta_{k}$ does not depend on the past states, actions and exogenous information, i.e., $\{S_{k'},a_{k'},W_{k'}\}_{k'\leq k-1}$. \par

Meanwhile, the reverse process of the proof is also true. This completes the proof of Theorem 1. \par

 \subsection{Proof of Theorem 2}
 We first prove the following Propostion 1.
\newtheorem{proposition}{Proposition}
\begin{proposition}
Given the SSDP in Definition 1 is also an MDP, we have
\begin{equation}
\mathrm{Pr}(W_{k}|S_{k},\tilde{S}_{k-1},a_{k-1})=\mathrm{Pr}(W_{k}|S_{k})
\end{equation}
\end{proposition}

Given the SSDP in Definition 1 is also an MDP, we have $\mathrm{Pr}(W_k| S_k, a_k) 
=\mathrm{Pr}(W_k| S_k, a_k, \{S_{k'},a_{k'}\}_{k'\leq k-1})$. Taking the expectation with regard to $a_k$ on both sides of the equation, we have
\begin{align}
\label{aeq10}
\mathrm{Pr}(W_k| S_k) 
&=\mathrm{Pr}(W_k| S_k, \{S_{k'},a_{k'}\}_{k'\leq k-1}) \IEEEnonumber \\
&=\mathrm{Pr}(W_k| S_k,S_{k-1},a_{k-1}) \IEEEnonumber \\
&=\mathrm{Pr}(W_k| S_k,S_{k-1},a_{k-1},W_{k-1}) \IEEEnonumber \\
\end{align}
\noindent where the last equation follows from $W_{k-1}=f^{\rm WS}(S_k,S_{k-1},a_{k-1})$ since $S_{k}=f^{\rm S}(S_{k-1},a_{k-1},W_{k-1})$, and the property of conditional probability $\mathrm{Pr}(X|g(Y),Y)=\mathrm{Pr}(X|Y)$. Thus, Proposition 1 is proved.

When the SSDP in Definition 1 is not an MDP, we can first convert it to an MDP by including the history in the state. In the following, we will only consider the case when the SSDP in Definition 1 is an MDP. In this case, the augmented-state SSDP is also an MDP having a state transition function
\begin{align}
\label{aeq7}
\tilde{S}_{k+1}&=
\begin{pmatrix}
S_{k+1}\\ W_{k+1}
\end{pmatrix}
\IEEEnonumber \\	
&=
\begin{pmatrix}
f^{S}(S_{k},a_{k},W_{k})\\ 
f^{W}(f^{S}(S_{k},a_{k},W_{k}),\tilde{\pi}({f^{S}(S_{k},a_{k},W_{k})}),\zeta_{k})
\end{pmatrix}\IEEEnonumber \\	
&=f^{\tilde{S}}(\tilde{S}_{k},a_{k},\zeta_{k}),
\end{align}
where the exogenous information $\zeta_{k}$ is an independent random variable with given distribution.\par

We first consider the finite horizon problem, where there are $K$ time steps indexed by $k\in\{0,1,\cdots,K-1\}$. \par 
Let $v_{k}^{*}(S_{k})$ and $\tilde{v}_{k}^{*}(\tilde{S}_{k})$ denote the value functions under the optimal policies $\pi^{*}$ and $\tilde{\pi}^{*}$ for the original SSDP and augmented-state SSDP, respectively. Define $\tilde{v}_{k}^{*}(S_{k})=\mathrm{E}_{W_{k}}[\tilde{v}_{k}^{*}(\tilde{S}_{k})|S_{k}]$.\par 

Note that $\tilde{J}^{*}=\mathrm{E}_{\tilde{S}_{0}}[\tilde{v}_{0}^{*}(\tilde{S}_{0})]=\mathrm{E}_{S_{0}}[\tilde{v}_{0}^{*}(S_{0})]$ and $J^{*}=\mathrm{E}_{S_{0}}[v_0^{*}(S_{0})]$. Therefore, in order to prove that $\tilde{J}^{*}\geq J^{*}$, it is sufficient to prove 
\begin{equation}
\label{aeq6}
\tilde{v}_{k}^{*}(S_{k})\geq v_{k}^{*}(S_{k}), \forall \ S_{k} \ \mathrm{and} \ k.
\end{equation} 

We will show \eqref{aeq6} by induction. For the last time step $K$, we have
\begin{equation}
\label{aeq1}
\tilde{v}_{K}^{*}(S_{K})= \mathrm{E}_{W_{K}}\Big[\max_{\tilde{\mu}_{K}(\tilde{S}_{K})}r\big(\tilde{S}_{K},\tilde{\mu}_{K}(\tilde{S}_{K})\big)|S_{K}\Big], \forall S_{K},
\end{equation}
\begin{equation}
\label{aeq2}
v_{K}^{*}(S_{K})= \max_{\mu_{K}(S_{K})}\mathrm{E}_{W_{K}}\Big[r\big(\tilde{S}_{K},\mu_{K}(S_{K})\big)|S_{K}\Big], \forall S_{K}.
\end{equation}
According to \eqref{aeq1} and \eqref{aeq2}, we have
\begin{equation}
\label{aeq10a}
\tilde{v}_{K}^{*}(S_{K})\geq v_{K}^{*}(S_{K}), \forall S_{K},
\end{equation} 
\noindent(since we generally have $\mathrm{E}[\max\{\cdot\}]\geq\max\{E[\cdot]\}$ according to Jensen's inequality). Therefore, the optimal action $\tilde{\pi}_{K}(\tilde{S}_{K})$ for the augmented-state SSDP problem is at least as good as that for the original SSDP $\pi_{K}(S_{K})$ at time step $K$. 

Assume that
\begin{equation}
\label{aeq9}
\tilde{v}_{k+1}^{*}(S_{k+1})\geq v_{k+1}^{*}(S_{k+1}), \forall  S_{k+1}.
\end{equation}

Consider the Bellman Equation for the original SSDP as 
\begin{align}
\label{aeq8}
&v_{k}^{*}(S_{k})=\max_{\pi_{k}(S_{k})}\Biggl\{\mathrm{E}_{W_{k}}\biggl[r\bigl(S_{k},\pi_{k}(S_{k}),W_{k}\bigr)+v_{k+1}^{*}\bigl(f^{S}(S_{k},\IEEEnonumber \\&\pi_{k}(S_{k}),W_{k})\bigr)\Big|S_{k},\pi_{k}(S_{k})\biggr]\Biggr\},
\end{align}

\noindent and consider the Bellman Equation for the augmented-state problem as 
\begin{align}
\label{aeq4}
&\tilde{v}_{k}^{*}(\tilde{S}_{k}) =\max_{\tilde{\pi}_{k}(\tilde{S}_{k})}\Biggl\{r\bigl(\tilde{S}_{k},\tilde{\pi}_{k}(\tilde{S}_{k})\bigr)+\mathrm{E}_{\tilde{S}_{k+1}}\biggl[\tilde{v}_{k+1}^{*}(S_{k+1},\IEEEnonumber \\&W_{k+1}) \Big|\tilde{S}_{k},\tilde{\pi}_{k}(\tilde{S}_{k})\biggr]\Biggr\}. 
\end{align}

Taking the expectation over $W_{k}$ conditioned on $S_{k}$ on both sides of \eqref{aeq4}, we have the following Bellman equation
\begin{align}
\label{aeq5}
&\tilde{v}_{k}^{*}(S_{k})=\mathrm{E}_{W_{k}}\bigl[\tilde{v}_{k}^{*}(\tilde{S}_{k})|S_{k}\bigr]\IEEEnonumber \\
&=\mathrm{E}_{W_{k}}\Biggl[\max_{\tilde{\pi}_{k}(\tilde{S}_{k})}\Biggl\{r\bigl(\tilde{S}_{k},\tilde{\pi}_{k}(\tilde{S}_{k})\bigr)+\mathrm{E}_{\tilde{S}_{k+1}}\biggl[\tilde{v}_{k+1}^{*}\Bigl(S_{k+1},W_{k+1}\Bigr) \IEEEnonumber \\
&\Big|\tilde{S}_{k},\tilde{\pi}_{k}(\tilde{S}_{k})\biggr]\Biggr\}\bigg|S_{k}\Biggr] \IEEEnonumber \\
&\stackrel{(a)}{\geq}\max_{\pi_{k}(S_{k})}\Biggl\{\mathrm{E}_{W_{k}}\Biggl[r\bigl(\tilde{S}_{k},\pi_{k}(S_{k})\bigr)+\mathrm{E}_{\tilde{S}_{k+1}}\biggl[\tilde{v}_{k+1}^{*}(S_{k+1},W_{k+1}) \IEEEnonumber \\
&\Big|\tilde{S}_{k},\pi_{k}(S_{k})\biggr]\bigg|S_{k})\Biggr]\Biggr\} \IEEEnonumber \\
&\stackrel{(b)}{=}\max_{\pi_{k}(S_{k})}\Biggl\{\mathrm{E}_{W_{k}}\Biggl[r\bigl(\tilde{S}_{k},\pi_{k}(S_{k})\bigr)+\mathrm{E}_{W_{k+1}}\Bigl[\tilde{v}_{k+1}^{*} \IEEEnonumber \\
&\bigl(f^{\rm S}(\tilde{S}_{k},\pi_{k}(S_{k})),W_{k+1}\bigr)\Big|\tilde{S}_{k},\pi_{k}(S_{k})\Bigr]\bigg|S_{k}\Biggr]\Biggr\},  \IEEEnonumber \\
&\stackrel{(c)}{=}\max_{\pi_{k}(S_{k})}\Biggl\{\mathrm{E}_{W_{k}}\Biggl[r\bigl(\tilde{S}_{k},\pi_{k}(S_{k})\bigr)+\mathrm{E}_{S_{k+1}}\Bigl[\mathrm{E}_{W_{k+1}}\Bigl[\tilde{v}_{k+1}^{*} \IEEEnonumber \\
&\bigl(f^{\rm S}(\tilde{S}_{k},\pi_{k}(S_{k})),W_{k+1}\bigr)|S_{k+1},\tilde{S}_{k},\pi_{k}(S_{k})\Bigr]\Bigr]\bigg|S_{k}\Biggr]\Biggr\},  \IEEEnonumber \\
&\stackrel{(d)}{=}\max_{\pi_{k}(S_{k})}\Biggl\{\mathrm{E}_{W_{k}}\Biggl[r\bigl(\tilde{S}_{k},\pi_{k}(S_{k})\bigr)+\mathrm{E}_{S_{k+1}}\Bigl[\mathrm{E}_{W_{k+1}}\Bigl[\tilde{v}_{k+1}^{*} \IEEEnonumber \\
&\bigl(f^{\rm S}(\tilde{S}_{k},\pi_{k}(S_{k})),W_{k+1}\bigr)|S_{k+1}\Bigr]\Bigr]\bigg|S_{k}\Biggr]\Biggr\},  \IEEEnonumber \\
&\stackrel{(e)}{=}\max_{\pi_{k}(S_{k})}\Biggl\{\mathrm{E}_{W_{k}}\Biggl[r\bigl(\tilde{S}_{k},\pi_{k}(S_{k})\bigr)+\mathrm{E}_{S_{k+1}}\Bigl[\tilde{v}_{k+1}^{*}(S_{k+1})\Bigr]\IEEEnonumber \\
&\bigg|S_{k}\Biggr]\Biggr\},  \IEEEnonumber \\
&\stackrel{(f)}{=}\max_{\pi_{k}(S_{k})}\Biggl\{\mathrm{E}_{W_{k}}\Biggl[r\bigl(\tilde{S}_{k},\pi_{k}(S_{k})\bigr)+\mathrm{E}_{S_{k+1}}\Bigl[\tilde{v}_{k+1}^{*}(S_{k+1})\Bigr]\IEEEnonumber \\
&\bigg|S_{k},\pi_{k}(S_{k})\Biggr]\Biggr\},  \IEEEnonumber \\
&\stackrel{(g)}{=}\max_{\pi_{k}(S_{k})}\Bigl\{\mathrm{E}_{W_{k}}\bigl[r\bigl(\tilde{S}_{k},\pi_{k}(S_{k})\bigr)+\tilde{v}_{k+1}^{*}(S_{k+1})|S_{k},\pi_{k}(S_{k})\bigr]\Bigr\}\IEEEnonumber \\
&\stackrel{(h)}{\geq}\max_{\pi_{k}(S_{k})}\Bigl\{\mathrm{E}_{W_{k}}\bigl[r\bigl(\tilde{S}_{k},\pi_{k}(S_{k})\bigr)+v_{k+1}^{*}(S_{k+1})|S_{k},\pi_{k}(S_{k})\bigr]\Bigr\}\IEEEnonumber \\
&\stackrel{(i)}{=}v_{k}^{*}(S_{k}),
\end{align}
\noindent where (a) follows by interchanging the expectation and maximization (since we generally have $\mathrm{E}[\max\{\cdot\}]\geq\max\{E[\cdot]\}$ according to Jensen's inequality); (b) follows from $\tilde{S}_{k+1}=\{S_{k+1},W_{k+1}\}$, $S_{k+1}=f^{\rm S}(\tilde{S}_{k},\pi_{k}(S_{k}))$, and the property of conditional probability $\mathrm{E}[g(X)|X]=g(X)$; (c) is due to the property of conditional probability $\mathrm{E}_{X}[\mathrm{E}[Y|X]]=\mathrm{E}[Y]$; (d) follows from Proposition 1, so we have $\mathrm{Pr}(W_{k+1}|S_{k+1},\tilde{S}_{k},\pi_{k}(S_{k}))=\mathrm{Pr}(W_{k+1}|S_{k+1})$; (e) follows from the definition of $\tilde{v}_{k+1}^{*}(S_{k+1})$; (f) holds because the expectation $\mathrm{E}_{W_{k}}[\cdot]$ is calculated for every action $\pi_{k}(S_{k})$ in the action space; (g) follows from $\mathrm{E}_{S_{k+1}}[\cdot|S_{k},\pi_{k}(S_{k})]=\mathrm{E}_{W_{k}}[\cdot|S_{k},\pi_{k}(S_{k})]$ since $S_{k+1}=f^{\rm S}(\tilde{S}_{k},\pi_{k}(S_{k}))$;
(h) follows from \eqref{aeq9}; and (i) follows from the Bellman equation for the original SSDP as given in \eqref{aeq8}. Thus \eqref{aeq6} is proved for all $k$ and the desired results are shown for the finite horizon problems. 

For the infinite horizon problems, the same proof above still holds when $K\rightarrow\infty$. Since the Bellman operator is a contraction mapping, the value functions will converge so we have
\begin{equation}
\label{aeq15}
\tilde{v}^{*}(s)\geq v^{*}(s), \forall s\in \mathcal{S},
\end{equation} 
\noindent which is similar to \eqref{aeq6} for the finite horizon problems. Thus, we have $\tilde{J}^{*}=\mathrm{E}_{\tilde{S}_{0}}[\tilde{v}^{*}(\tilde{S}_{0})]=\mathrm{E}_{S_{0}}[\tilde{v}_{0}^{*}(S_{0})]\geq J^{*}=\mathrm{E}_{S_{0}}[v^{*}(S_{0})]$. This completes the proof of Theorem 2.

	\bibliographystyle{IEEEtran}
	\bibliography{IEEEabrv,VoI}
	
\end{document}